\documentclass[letterpaper]{article}
\usepackage[preprint]{aaai2027}
\usepackage[hyphens]{url}
\usepackage{graphicx}
\usepackage{natbib}
\usepackage{caption}
\usepackage{booktabs}
\usepackage{amsmath}
\usepackage{amssymb}
\usepackage[table]{xcolor}
\usepackage{array}
\usepackage{multirow}
\usepackage[most]{tcolorbox}
\usepackage{tabularray}  

\definecolor{prov}{gray}{0.55}

\newcommand{\FIN}[2]{#1\,{\tiny$\pm$#2}}
\newcommand{\ORCA}{\textbf{ORCA}}

\title{ORCA: ORgan-Centroid Aggregation for\\ Training-Free 3D CT Visual Token Compression}
\author{
Renjie Liang\textsuperscript{\rm 1},
Zijian Xu\textsuperscript{\rm 1},
Jinqian Pan\textsuperscript{\rm 1},
Chengkun Sun\textsuperscript{\rm 1}, \\
Zhengkang Fan\textsuperscript{\rm 1}, 
Shawn Li\textsuperscript{\rm 2},
You Qin\textsuperscript{\rm 3},
Mei Liu\textsuperscript{\rm 1},
Jie Xu\textsuperscript{\rm 1}
}
\affiliations{
\textsuperscript{\rm 1}University of Florida, Gainesville, FL, USA\\
\textsuperscript{\rm 2}University of Southern California, Los Angeles, CA, USA\\
\textsuperscript{\rm 3}National University of Singapore, Singapore
}

\begin{document}
\maketitle

\begin{abstract}

\begin{quote}
A 3D CT scan entering a vision-language model produces a long sequence of visual tokens, often thousands to tens of thousands per volume, and this sequence must be compressed before a language model can consume it. Token compression is well studied in general vision, but little of it targets 3D CT specifically. A common baseline is grid average, which pools regular grid cells and can blend distinct anatomy, lesion, and air into one token. We present \textbf{ORCA} (ORgan-Centroid Aggregation), a token compressor for 3D CT. It merges adjacent tokens with organ guidance and adds a sinusoidal encoding of each region's centroid to preserve spatial layout. This preserves the anatomical information a downstream model needs. ORCA is training-free and plug-and-play, producing an adjustable token set without any model change or text query. We evaluate it across two datasets (CT-RATE and Merlin) and five encoders. The evaluation spans two task types: attribute prediction over five families (size, density, location, texture, and disease) and text generation (visual question answering and report generation). At matched token budgets, ORCA improves consistently over existing compression methods. It shrinks the visual context $64\times$ and its KV-cache $50\times$, and is $31\times$ faster to process each volume. Code released at \url{https://github.com/renjie-liang/ORCA-3DCT}.\end{quote}

\end{abstract}

\noindent\textbf{Keywords:} Token compression, 3D CT, vision-language models, training-free, medical report generation
\section{Introduction}

3D computed tomography (CT) is used at national clinical scale and is becoming a practical input to medical vision--language systems. A CT-volume analysis spanning 2{,}398 U.S. radiology practices shows how broadly CT is used in routine care \cite{davenport2021ctvolumes}; a recent workload study of 46.4 million imaging examinations from 167 facilities found that the highest-volume radiologists read 30.6\% more examinations per day and worked 19.7\% more clinical days per quarter by 2024 \cite{zamani2026us}. At the same time, richer 3D CT datasets and models now support abnormality detection \cite{hamamci2024ctrate,blankemeier2026merlin}, visual question answering \cite{wu2025generalist}, and report generation from volumetric studies \cite{hamamci2024ct2rep}. A common 3D CT vision--language pipeline maps a volume into a grid of visual tokens, then feeds them through a projector to an LLM or another downstream model. Because a CT scan contains dense anatomical information across many slices, this grid is often large. For example, COLIPRI \cite{wald2025colipri} emits a $24^3$ grid, or 13{,}824 visual tokens. These tokens must be compressed before downstream use. Compression therefore changes more than sequence length: it determines which anatomical evidence remains available to the downstream model.

Existing methods for visual token compression address this bottleneck in three ways \cite{shao2025tokens}. The \emph{Grid average} baseline pools visual tokens over regular 3D grid cells. It is simple and strong, but blind to anatomy: one pooled cell can mix organ tissue, lesions, vessels, and air. For instance, a small nodule or a calcified plaque occupies only a few tokens, and a grid cell that averages it into surrounding lung or muscle erases the very finding a downstream task must read. Pruning keeps tokens that appear important and discards the rest \cite{chen2024fastv,yang2024visionzip,liu2026medpruner}. Which is efficient, but it turns compression into a hard decision: evidence from a removed region cannot be recovered later, and many pruning rules require attention scores or a text query. A third route builds compact anatomy representations into the encoder itself \cite{shui2025fvlm,cao2025visdboost}. These models are powerful, but the compression rule is tied to the architecture and training objective, so it cannot serve as a standalone compressor at the encoder output with a different token budget. This leaves a gap: we must compress the visual tokens a 3D CT encoder produces while preserving anatomical evidence later tasks may need.

We introduce \textbf{ORCA} (ORgan-Centroid Aggregation), a training-free compressor for 3D CT visual tokens. ORCA aggregates neighboring tokens into connected regions, using visual similarity and organ guidance to avoid mixing unrelated anatomy. The guidance does not force tokens to be pooled by predefined organ masks. ORCA stops at the target token budget, and each region becomes one compressed visual token. It also adds a centroid position encoding after aggregation. This restores spatial layout for merged regions whose original grid order no longer carries reliable position. It requires no model surgery, attention hooks, additional supervision, or changes to the downstream training recipe. Our key contributions are as follows:
\begin{itemize}
\item We propose \textbf{ORCA}, a training-free compressor for 3D CT visual tokens that aggregates spatially connected, feature-similar regions with organ guidance and restores region position with sinusoidal centroid encoding.
\item We evaluate ORCA on CT-RATE chest CT and Merlin abdomen CT across five encoders, five attribute families, and text generation. It preserves more information than the other compressors, consistently across every encoder and attribute family, and at $64\times$ compression it stays within a small margin of the uncompressed tokens.
\item We show the encoder and the compressor play separate roles: the encoder sets how much information the tokens carry, and the compressor governs how much of it survives to the downstream model. 
\end{itemize}

\begin{figure*}[!tp]
\centering
\centering
\includegraphics[width=1.0\textwidth]{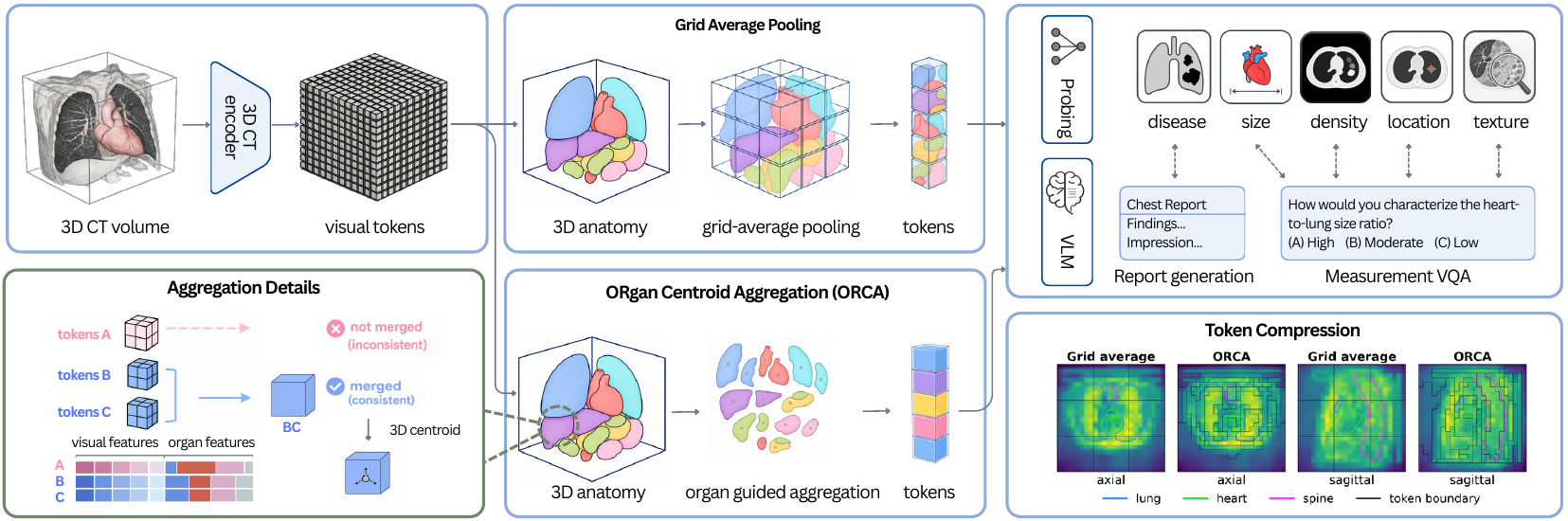}
\caption{ORCA overview. A frozen 3D encoder turns a CT volume into dense visual tokens. Grid average pools them over fixed cells; ORCA instead aggregates spatially connected tokens into organ-guided regions and appends each region's 3D centroid. Bottom-right: ORCA's token boundaries track anatomy under compression.}
\label{fig:method_overview}
\vspace{-10pt}
\end{figure*}
\section{Related Work}

\textbf{3D CT vision--language models.} Recent work has made volumetric CT a practical input to medical vision--language systems. CT-RATE introduced large-scale CT-volume and radiology-report data and supports CT-CLIP and CT-CHAT style modeling \cite{hamamci2024ctrate}. CT-GRAPH studies anatomy-guided report generation from 3D CT \cite{kalisch2025ctgraph}, while Merlin extends CT foundation modeling to abdominal CT and multimodal clinical context \cite{blankemeier2026merlin}. Generalist radiology and 3D medical VLMs further broaden the task suite to visual question answering, report generation, retrieval, localization, and segmentation \cite{wu2025generalist,bai2024m3d,xin2025med3dvlm}. COLIPRI focuses on training a stronger 3D CT encoder through language--image pretraining \cite{wald2025colipri}, while BTB3D learns compact volumetric tokens through reconstruction-oriented encoding of full 3D CT volumes \cite{hamamci2025btb3d}. These models establish the need for strong 3D visual tokens, but they treat compression as part of a fixed model pipeline. ORCA instead studies compression at the encoder-output interface and asks which visual evidence remains available under a target token budget.

\textbf{Visual token compression.} Visual token compression has been studied for vision transformers and multimodal LLMs. Shao et al. categorize multimodal token compression by modality and by mechanism, including transformation-based, similarity-based, attention-based, and query-based approaches \cite{shao2025tokens}. In visual models, these mechanisms appear as pooling or downsampling, token merging, pruning, and query-based resampling. Concrete methods realize these mechanisms: ToMe merges tokens, DivPrune selects diverse ones, FastV prunes by in-LLM attention, LLaVA-PruMerge and VisionZip combine selection with merging, and TokenPacker learns a compact resampler \cite{bolya2023tome,alvar2025divprune,chen2024fastv,shang2024llavaprumerge,yang2024visionzip,li2024tokenpacker}. Recent analyses caution that token-reduction gains depend on evaluation design: some pruning rules underperform naive random token selection, and simple image downsampling can outperform advanced compressors on common benchmarks \cite{wen2025rightproblem,liao2025vtcbench}. This matters for 3D CT because Grid average is a common baseline: it preserves the field of view, is easy to implement, and is compatible with downstream models. Its weakness is not simplicity, but anatomy blindness.

\textbf{Token-efficient CT systems.} CT-specific systems address token efficiency by changing the architecture or task pipeline. MedPruner shortens 3D medical visual sequences through inter-slice filtering followed by attention-based token pruning \cite{liu2026medpruner}. Photon uses instruction-conditioned token scheduling with surrogate gradient propagation for 3D medical VQA \cite{fang2026photon}. MedRegion-CT combines a region-based SlowFast tokenizer, pseudo-mask guidance, and structured lesion prompts for CT report generation \cite{kyung2025medregionct}. CT-GRAPH uses anatomical masks to extract global and organ-level features, then refines them in a hierarchy from organs to anatomical systems and global context before report generation \cite{kalisch2025ctgraph}. Ker-VLJEPA uses zone-constrained cross-attention to compress slice embeddings into spatially grounded tokens in a thoracic CT report-generation framework \cite{bumgardner2026kervljepa}. A related line of anatomy-grounded pretraining methods, including CT-GLIP, fVLM, and ViSD-Boost, shows that organ-level alignment and anatomy-level representations can improve 3D CT understanding \cite{lin2024ctglip,shui2025fvlm,cao2025visdboost}. These systems show that anatomy and token efficiency matter. They also solve a different problem from ours: their compression is tied to a particular model, task, or region definition, so none is a drop-in compressor at a matched budget. ORCA instead targets the simpler interface after a 3D encoder has produced a dense token grid, replacing Grid average with an adjustable compressor that keeps merged regions connected in 3D.

\section{Method}

\subsection{Problem setup}
A 3D CT encoder converts an input volume into a regular 3D grid of visual tokens. We denote the token set by $X=\{x_i\}_{i=1}^{M}$, where $x_i\in\mathbb{R}^{D}$ is the feature vector for token $i$, and $q_i\in[0,1]^3$ is its normalized 3D grid coordinate. The compressor reduces these $M$ input tokens to $B$ output tokens, denoted $Z=\{z_j\}_{j=1}^{B}$, where $B<M$. These $B$ tokens are then passed to the multimodal projector and LLM. In all compressor comparisons, only the compressor changes. For text generation experiments, the encoder, projector architecture, LLM, training data, and training schedule are fixed. For probing experiments, the readout architecture, training split, and evaluation metric are fixed.

\subsection{ORCA: ORgan-Centroid Aggregation}
Figure~\ref{fig:method_overview} summarizes ORCA. ORCA compresses a dense 3D token grid into $B$ connected regions. It starts from one region per token and repeatedly merges neighboring regions using visual similarity and soft organ-mask guidance. After merging, ORCA turns each region into one output token by averaging the visual embeddings inside the region and appending a sinusoidal encoding of the region's 3D centroid.

\subsubsection{Connected region aggregation}
ORCA uses Ward linkage \cite{ward1963hierarchical} as its merge criterion, but applies it to 3D CT token compression rather than unconstrained clustering. Merges are restricted to standard 6-neighbor connectivity in the token grid, where regions are adjacent if they share a face. The compressor outputs $B$ non-overlapping regions that cover the original token grid,
{\scriptsize
\[
\mathcal{P}_B=\{R_1,\ldots,R_B\},\;
\bigcup_{j=1}^{B}R_j=\{1,\ldots,M\},\;
R_a\cap R_b=\emptyset\ (a\ne b).
\]
}
Each region must be spatially connected in the original 3D grid, which prevents one output token from mixing disconnected parts of the scan.

ORCA initializes each token as its own region. At each step, it considers only adjacent region pairs. Let $\tilde{x}_i$ be the feature used to score merges. The next subsection defines how organ guidance constructs it. For a region $R$, let $\tilde{\mu}_R=|R|^{-1}\sum_{i\in R}\tilde{x}_i$ be its mean in this merge-feature space. For two adjacent regions $R_a$ and $R_b$, ORCA computes the Ward increase in within-region distortion,
{\footnotesize
\[
\Delta(R_a,R_b)=
\frac{|R_a||R_b|}{|R_a|+|R_b|}
\left\|\tilde{\mu}_a-\tilde{\mu}_b\right\|_2^2 .
\]
}
It merges the pair with the smallest $\Delta$ and repeats until exactly $B$ connected regions remain.

\subsubsection{Organ guidance}
ORCA uses organ masks deciding which regions to merge. In our experiments, these masks are produced by TotalSegmentator~\cite{totalsegmentator}. Two neighboring regions are favored for merging when both their visual embeddings and their organ coverage are similar. The organ information remains soft: it acts only on the merge cost. Regions may span several organs and need not follow the predefined organ masks.

For each token $i$, let $m_i\in[0,1]^K$ denote its organ-coverage vector. The entry $m_{i,k}$ is the proportion of voxels assigned to organ group $k$ within the 3D patch represented by token $i$. ORCA makes this organ information comparable to the visual embedding before using it for merging. It first computes two scale terms over all input tokens in the current volume,
{\footnotesize
\[
S_x=\sum_{d=1}^{D}\operatorname{Var}_i(x_{i,d}),\qquad
S_m=\sum_{k=1}^{K}\operatorname{Var}_i(m_{i,k})+\epsilon ,
\]
}
where $S_x$ and $S_m$ are the total variances of the visual embedding and organ-coverage dimensions, and $\epsilon$ prevents division by zero. ORCA then scales the organ vector and concatenates it with the visual embedding to form the merge feature,
{\footnotesize
\[
\alpha=\sqrt{\lambda S_x/S_m},\qquad \tilde{x}_i = [x_i;\alpha m_i].
\]
}
The parameter $\lambda$ controls how strongly organ coverage affects the merge. The augmented feature is used only to build the merge tree. The visual feature part of each output token is the mean of the original visual embeddings:
{\footnotesize
\[
\mu_j = \frac{1}{|R_j|}\sum_{i\in R_j} x_i .
\]
}
Organ coverage therefore shapes which tokens merge, not what they contain: each output token is a mean of original embeddings, with every input token contributing to exactly one output.

\subsubsection{Centroid position encoding}
After ORCA merges neighboring token regions, the output regions no longer form a regular grid. Their order in the output sequence therefore does not reliably indicate where they came from in the scan. ORCA records this location explicitly. For each region $R_j$, it computes the normalized 3D centroid
{\footnotesize
\[
c_j=\frac{1}{|R_j|}\sum_{i\in R_j}q_i .
\]
}
It then appends a sinusoidal encoding of this centroid~\cite{vaswani2017attention},
{\footnotesize
\[
\phi(c_j)=
\big[\sin(2^{k}\pi c_{j,a}),\,\cos(2^{k}\pi c_{j,a})\big]_
{a\in\{x,y,z\},\,k=0}^{F-1}.
\]
}
For $F{=}4$ frequencies over three axes, $\phi$ adds 24 position dimensions. These dimensions are normalized per dimension and scaled by a factor $s$ to match the typical magnitude of the visual features before concatenation,
{\footnotesize
\[
z_j=[\mu_j;\,s\,\widehat{\phi}(c_j)] .
\]
}
Each output token therefore carries its own location in its value, so regions of any shape or size stay localizable.

\section{Experiments}

\subsection{Experimental setup}
\subsubsection{Datasets and encoders}
We evaluate ORCA on CT-RATE and Merlin. CT-RATE is a non-contrast chest CT dataset with paired radiology reports and abnormality labels \cite{hamamci2024ctrate}. Merlin is a portal-venous abdomen CT corpus with shipped reports and 30 clinical findings \cite{blankemeier2026merlin}. Table~\ref{tab:encoders} summarizes the raw encoder-token grids and dimensions.

\begin{table}[!tbp]
\centering\scriptsize
\setlength{\tabcolsep}{3pt}
\begin{tabular}{@{}lllcc@{}}
\toprule
corpus & encoder & pretraining & grid & $D$ \\
\midrule
CT-RATE & COLIPRI~\cite{wald2025colipri} & language--image & $24{\times}24{\times}24$ & 768 \\
CT-RATE & CT-CLIP~\cite{hamamci2024ctrate} & language--image & $24{\times}24{\times}24$ & 512 \\
CT-RATE & BTB3D~\cite{hamamci2025btb3d} & reconstruction & $31{\times}32{\times}32$ & 18 \\
Merlin & SuPreM~\cite{suprem} & segmentation & $12{\times}12{\times}12$ & 192 \\
Merlin & SegVol~\cite{segvol} & segmentation & $8{\times}16{\times}16$ & 768 \\
\bottomrule
\end{tabular}
\caption{Encoders used in the experiments. The pretraining column is the encoder's training objective; $D$ is the raw token dimension before compression.}
\label{tab:encoders}
\end{table}

\subsubsection{Attribute probing}
Attribute probing provides a representation-level readout of what information remains in compressed visual tokens \cite{alain2016understanding,belinkov2022probing}, and it is a task in its own right, since attribute prediction is how structured findings are populated in practice. In our protocol, a lightweight readout is trained directly on the compressed tokens before the projector and language model, so the measurement does not involve LLM fine-tuning, prompting, or text decoding. We evaluate this readout using the probing benchmark of \citet{liang2026cheap}. The benchmark organizes abnormality labels and measurements from CT volumes and segmentation masks into five families: disease, size, density, location, and texture. In this benchmark, an \emph{attribute} is an individual target within one of these families. Table~\ref{tab:probe_families} summarizes the families, example attributes, metrics, and corresponding generation tasks. The supplementary material reports the full attribute list and per-attribute scores.

\subsubsection{Report generation}
Report generation evaluates compressed visual tokens in a longer-form language task. In the CT-RATE setting, the model generates a radiology report that is compared with the paired reference report and CT-RATE abnormality labels~\cite{hamamci2024ctrate}. We report lexical overlap metrics, including BLEU-1, BLEU-4, and ROUGE-L \cite{papineni2002bleu,lin2004rouge}, and clinical-efficacy metrics: CE F1 over the abnormality labels a radiology labeler~\cite{radbert} assigns to each generated report, together with CRG and GREEN \cite{hamamci2025crg,ostmeier2024green}. Report generation is therefore the generation task associated with the disease family: abnormality labels are read from compressed tokens in probing and from generated text in clinical metrics.

\begin{table}[!t]
\centering\scriptsize
\setlength{\tabcolsep}{4pt}
\begin{tabular*}{\columnwidth}{@{\extracolsep{\fill}}lllll}
\toprule
family & attributes & probe task & metric & generation task \\
\midrule
disease & 18 abnormalities & multi-label cls. & AUROC & report gen. \\
size & heart/aorta/IVC/lung & regression & $R^2$ & measurement VQA \\
density & lung/spine/aorta HU & regression & $R^2$ & measurement VQA \\
location & organ position & regression & $R^2$ & measurement VQA \\
texture & lung texture & regression & $R^2$ & measurement VQA \\
\bottomrule
\end{tabular*}
\caption{Attribute-probing families and their corresponding generation tasks.}
\label{tab:probe_families}
\end{table}

\subsubsection{Measurement VQA}
Measurement VQA is the generation task associated with the four remaining families: size, density, location, and texture. Its gold answers are direct anatomical measurements. We use the measurement VQA questions of \citet{liang2026cheap}; Appendix~\ref{app:vqabench} summarizes the question types and gives examples. Continuous attributes become three-choice tertile or two-choice clinical-threshold questions, and we report exact match accuracy for each family. Because each question targets a specific organ's attribute, accuracy measures how much precise, localized detail a compressor preserves rather than a coarse whole-volume impression.

\begin{table}[!bp]
\centering\scriptsize
\setlength{\tabcolsep}{3pt}
\begin{tabular}{llccccc}
\toprule
\multirow{2}{*}{method} & \multirow{2}{*}{$B$} & disease & size & density & location & texture \\
 & & (AUROC) & \multicolumn{4}{c}{($R^2$)} \\
\midrule
Uncompressed & 13{,}824 & 0.848 & 0.727 & 0.915 & 0.252 & 0.811 \\
\midrule
\multirow{2}{*}{Grid average}
 & 27 & 0.850 & 0.676 & 0.806 & 0.265 & 0.683 \\
 & 216 & 0.851 & 0.681 & 0.865 & 0.247 & 0.760 \\
\midrule
Slice pooling & 24 & 0.850 & 0.649 & 0.752 & 0.172 & 0.646 \\
\midrule
\multirow{2}{*}{\shortstack[l]{DivPrune\\ \cite{alvar2025divprune}}}
 & 27 & 0.850 & 0.652 & 0.854 & 0.228 & 0.741 \\
 & 216 & 0.851 & 0.681 & 0.887 & 0.260 & 0.753 \\
\midrule
\multirow{2}{*}{\shortstack[l]{MedPruner-DINS\\ \cite{liu2026medpruner}}}
 & 27 & 0.847 & 0.635 & 0.848 & 0.175 & 0.706 \\
 & 216 & 0.851 & 0.673 & 0.903 & 0.233 & 0.765 \\
\midrule
\multirow{2}{*}{\shortstack[l]{ToMe\\ \cite{bolya2023tome}}}
 & 27 & 0.844 & 0.617 & 0.786 & 0.172 & 0.647 \\
 & 216 & 0.848 & 0.636 & 0.845 & 0.220 & 0.714 \\
\midrule
\shortstack[l]{MedRegion-CT\\ \cite{kyung2025medregionct}} & $\bar N{=}549$ & 0.850 & 0.694 & 0.902 & 0.271 & 0.810 \\
\midrule
\multirow{2}{*}{\ORCA{}}
 & 27 & \textbf{0.851} & \textbf{0.691} & \textbf{0.896} & \textbf{0.622} & \textbf{0.775} \\
 & 216 & \textbf{0.852} & \textbf{0.720} & \textbf{0.913} & \textbf{0.677} & \textbf{0.816} \\
\bottomrule
\end{tabular}
\caption{CT-RATE probing on COLIPRI. Bold marks the best per budget, excluding the uncompressed reference. Cells are three-seed means; standard deviations are ${\le}0.03$. MedRegion-CT is reported at its mean count, $\bar N{=}549$.}
\label{tab:probe_colipri}
\end{table}

\subsubsection{Baselines}
Each baseline is an alternative compressor in ORCA's slot, with the rest of the pipeline fixed as above. Grid average is a simple baseline that averages fixed grid cells, and slice pooling averages whole slices. DivPrune~\cite{alvar2025divprune} selects a diverse token subset and drops the rest. MedPruner-DINS~\cite{liu2026medpruner} keeps high-attention tokens and folds lower-attention tokens into them. ToMe~\cite{bolya2023tome} merges tokens by feature similarity, without the 3D connectivity or anatomical guidance ORCA uses. MedRegion-CT pooling~\cite{kyung2025medregionct} averages tokens within each organ mask, the same-domain organ-pooling baseline. We also report an uncompressed-grid reference. Some of these baselines were designed for 2D-slice-stack encoders, so we adapt each one faithfully to our 3D-native interface; Appendix~\ref{app:baseline_adapt} gives the exact realization of every baseline.

\subsubsection{Matched-budget protocol}
To attribute differences to \emph{which} tokens survive rather than how many, we fix the token budget whenever a method exposes an adjustable target count. Some baselines instead have a count fixed by construction. We report those methods at their natural operating point. MedRegion-CT pooling is the main such case because its count is fixed by anatomy.

\begin{table}[!tbp]
\centering\scriptsize
\setlength{\tabcolsep}{3pt}
\begin{tabular*}{\columnwidth}{@{\extracolsep{\fill}}llccccc}
\toprule
method & $B$ & disease & size & density & location & texture \\
\midrule
\multicolumn{7}{l}{\emph{CT-CLIP} } \\
\multirow{2}{*}{Grid average} & 27 & \textbf{0.741} & 0.584 & \textbf{0.560} & 0.516 & \textbf{0.557} \\
 & 216 & 0.743 & 0.612 & 0.570 & 0.529 & \textbf{0.566} \\
\multirow{2}{*}{ToMe} & 27 & 0.665 & 0.246 & 0.306 & 0.155 & 0.350 \\
 & 216 & 0.715 & 0.422 & 0.472 & 0.256 & 0.499 \\
\multirow{2}{*}{\ORCA{}} & 27 & 0.738 & \textbf{0.615} & 0.548 & \textbf{0.649} & 0.542 \\
 & 216 & \textbf{0.748} & \textbf{0.649} & \textbf{0.586} & \textbf{0.645} & 0.564 \\
\midrule
\multicolumn{7}{l}{\emph{BTB3D} } \\
\multirow{2}{*}{Grid average} & 27 & 0.616 & 0.171 & 0.194 & 0.080 & 0.179 \\
 & 216 & 0.608 & 0.179 & 0.211 & 0.064 & 0.180 \\
\multirow{2}{*}{ToMe} & 27 & 0.588 & 0.100 & 0.123 & 0.039 & 0.119 \\
 & 216 & 0.599 & 0.126 & 0.157 & 0.046 & 0.144 \\
\multirow{2}{*}{\ORCA{}} & 27 & \textbf{0.672} & \textbf{0.420} & \textbf{0.345} & \textbf{0.564} & \textbf{0.361} \\
 & 216 & \textbf{0.679} & \textbf{0.440} & \textbf{0.405} & \textbf{0.434} & \textbf{0.401} \\
\bottomrule
\end{tabular*}
\caption{CT-RATE probing on CT-CLIP and BTB3D. Disease is macro-AUROC; other families are $R^2$. Cells are means over three seeds with standard deviations ${\le}0.03$.}
\vspace{-10pt}
\label{tab:probe_ctclip_btb3d}
\end{table}

\subsection{Probing results}
\label{sec:probing}

\subsubsection{CT-RATE}
Tables~\ref{tab:probe_colipri} and~\ref{tab:probe_ctclip_btb3d} report the three CT-RATE encoders. On COLIPRI, ORCA is level with the best baselines on size, density, and texture and far ahead on location: $0.677$ at $B{=}216$ against $0.247$ for Grid average and $0.271$ for MedRegion-CT pooling, and it keeps almost all of that lead down to $B{=}27$. The location gain comes from recording where each region sits: ORCA appends a sinusoidal encoding of the region centroid~\cite{tancik2020fourier,vaswani2017attention}, which a model reads far more easily than bare coordinates.

The other two encoders place ORCA in a wider frame. They are trained for different objectives: BTB3D for pixel reconstruction, CT-CLIP for report--image contrastive alignment, and COLIPRI for that alignment plus report generation and masked image modeling. Objectives tied to language leave more of the probed content in the tokens than pixel reconstruction does, so on the attributes we measure COLIPRI $>$ CT-CLIP $>$ BTB3D. Two factors then separate cleanly: \emph{the encoder fixes the ceiling, the compressor decides how much of it survives.} COLIPRI with ORCA is the best pair on nearly every family. What makes ORCA general is that it has no failure mode: across encoders and all five attributes it stays at or near the best, whereas the alternatives each fall short somewhere, most visibly on location.

\begin{table}[!tbp]
\centering\scriptsize
\setlength{\tabcolsep}{4pt}
\begin{tabular}{lllcccc}
\toprule
encoder & method & $B$ & disease & size & density & location \\
\midrule
\multirow{7}{*}{SuPreM}
 & Uncompressed & 1{,}728 & 0.792 & 0.510 & 0.749 & 0.431 \\
\cmidrule(l){2-7}
 & Grid average & 216 & 0.787 & 0.462 & 0.681 & 0.395 \\
 & ToMe & 216 & 0.777 & 0.375 & 0.649 & 0.300 \\
 & \ORCA{} & 216 & \textbf{0.795} & \textbf{0.544} & \textbf{0.795} & \textbf{0.600} \\
\cmidrule(l){2-7}
 & Grid average & 64 & 0.788 & 0.441 & 0.661 & 0.399 \\
 & ToMe & 64 & 0.768 & 0.289 & 0.584 & 0.235 \\
 & \ORCA{} & 64 & \textbf{0.788} & \textbf{0.536} & \textbf{0.792} & \textbf{0.602} \\
\midrule
\multirow{9}{*}{SegVol}
 & Uncompressed & 2{,}048 & 0.755 & 0.481 & 0.683 & 0.467 \\
\cmidrule(l){2-7}
 & Grid average & 256 & 0.746 & 0.466 & 0.626 & 0.378 \\
 & MedPruner-DINS & 256 & \textbf{0.765} & 0.566 & 0.667 & 0.560 \\
 & ToMe & 256 & 0.748 & 0.345 & 0.629 & 0.436 \\
 & \ORCA{} & 256 & 0.761 & \textbf{0.588} & \textbf{0.688} & \textbf{0.596} \\
\cmidrule(l){2-7}
 & Grid average & 32 & 0.738 & 0.338 & 0.556 & 0.322 \\
 & MedPruner-DINS & 32 & \textbf{0.758} & 0.460 & 0.597 & 0.432 \\
 & ToMe & 32 & 0.738 & 0.234 & 0.552 & 0.289 \\
 & \ORCA{} & 32 & 0.754 & \textbf{0.580} & \textbf{0.696} & \textbf{0.656} \\
\bottomrule
\end{tabular}
\caption{Merlin probing, means over three seeds. SegVol probing is noisier, sd up to $0.05$, while SuPreM is stable, sd ${\le}0.02$. Disease is macro-AUROC; size, density, and location are $R^2$. MedPruner-DINS runs on SegVol only; SuPreM's windowed attention exposes no global token saliency.}
\label{tab:merlin}
\end{table}

\begin{figure}[!b]
\centering

\includegraphics[width=\linewidth]{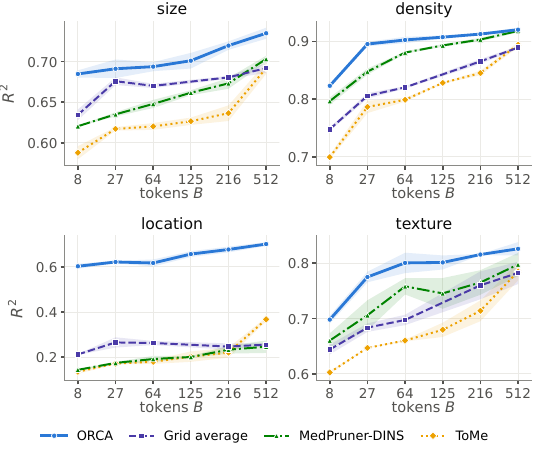}
\vspace{-10pt}

\caption{Budget curve on COLIPRI. Probe $R^2$ versus token budget. Each measured budget is an equal slot; the axis is not linear in $B$. Bands are $\pm1$ sd over three seeds. Grid average is ratio-based and is absent at the non-cubic budget $125$.}
\label{fig:budget_colipri}
\end{figure}

\begin{table*}[!tbp]
\centering\scriptsize
\setlength{\tabcolsep}{2.6pt}
\begin{tabular*}{0.85\textwidth}{@{\extracolsep{\fill}}ll cccccc cccc}
\toprule
 & & \multicolumn{6}{c}{report generation} & \multicolumn{4}{c}{Measurement VQA (accuracy)} \\
\cmidrule(lr){3-8}\cmidrule(lr){9-12}
method & $B$ & BLEU-1 & BLEU-4 & ROUGE-L & CE-F1 & CRG & GREEN
       & size & density & location & texture \\
\midrule
Noise tokens & 8 & 0.441 & 0.191 & 0.278 & 0.306 & 0.366 & 0.226 &  0.340 & 0.342 & 0.337 & 0.357 \\
\midrule
\multirow{2}{*}{Grid average}
 & 27  & \textbf{0.472} & \textbf{0.206} & 0.292 & 0.470 & 0.437 & 0.259 & 0.759 & 0.768 & 0.601 & 0.783 \\
 & 216 & 0.445 & 0.191 & 0.293 & 0.475 & 0.438 & 0.319 & 0.766 & 0.807 & 0.649 & 0.795 \\
\midrule
\multirow{2}{*}{MedPruner-DINS}
 & 27  & 0.458 & 0.195 & 0.293 & \textbf{0.482} & \textbf{0.442} & 0.288 & 0.737 & 0.800 & 0.494 & 0.811 \\
 & 216 & \textbf{0.454} & \textbf{0.196} & 0.297 & \textbf{0.483} & \textbf{0.444} & 0.317 & 0.759 & 0.837 & 0.527 & 0.830 \\
\midrule
\multirow{2}{*}{\ORCA{}}
 & 27  & 0.442 & 0.190 & \textbf{0.294} & 0.477 & \textbf{0.442} & \textbf{0.313} & \textbf{0.760} & \textbf{0.838} & \textbf{0.677} & \textbf{0.832} \\
 & 216 & 0.446 & \textbf{0.196} & \textbf{0.299} & 0.470 & 0.436 & \textbf{0.329} & \textbf{0.778} & \textbf{0.847} & \textbf{0.721} & \textbf{0.838} \\
\bottomrule
\end{tabular*}
\caption{Text generation results on CT-RATE, single seed. The Noise row replaces the visual tokens with Gaussian noise, giving a language-prior floor.}
\label{tab:textgen}
\end{table*}

\subsubsection{Merlin}
Table~\ref{tab:merlin} reports probing on Merlin with SuPreM and SegVol. Merlin has no texture family because the texture targets are lung-based. The pattern is the same: ORCA leads on size, density, and location at both budgets, most strongly on location, and disease is saturated and near-tied across methods. What makes this a real test is how much changes: Merlin is a different anatomy (abdomen) and acquisition (portal-venous), and SuPreM and SegVol are a third kind of encoder, segmentation-pretrained rather than language-supervised. That ORCA's advantage survives all of this is the strongest sign it is not tied to one encoder family or dataset. Its location lead even holds at the tightest budgets.

\paragraph{Budget curves.} Figure~\ref{fig:budget_colipri} sweeps the full budget range on COLIPRI. Every curve rises with the budget, since more tokens carry more of the attribute information, yet ORCA keeps its lead across the whole sweep and its location margin never closes. Disease is omitted because it stays near-saturated at every budget: a global finding that survives even heavy compression, so its curve is flat.

\subsection{Text generation results}
\label{sec:textgen}
We fine-tune the language model on each compressed cell for visual question answering (VQA) and report generation (Table~\ref{tab:textgen}), asking whether a compressor's advantage carries over to LLM generation tasks. On VQA, ORCA leads every family at both budgets, and the gap is widest on location: at $B{=}216$ ORCA reaches $0.721$ against $0.649$ for Grid average and $0.527$ for MedPruner-DINS. Appendix~\ref{app:vqabudget} reports additional budgets and Appendix~\ref{app:vqadyn} the training curves.

Report generation's three metric families separate methods to very different degrees. GREEN, the most comprehensive clinical score, separates them: ORCA leads at both budgets, widest at $B{=}27$ ($0.313$ against $0.288$ for MedPruner-DINS and $0.259$ for Grid average), while a noise-token control collapses to $0.226$, confirming the signal comes from the compressed tokens. Because GREEN grades many findings rather than one disease label, this ordering matches the overall probing and VQA picture. The coarse CE-F1 label barely separates compressors and at times inverts them, within $0.02$ across methods and budgets. This too matches probing, where the disease family is saturated, so a whole-volume judgment that survives heavy pooling cannot resolve compressors and is the more fragile signal in the longer generation pipeline. Lexical overlap carries the least: noise tokens still score BLEU-1 near $0.44$, indistinguishable from real compressors, since it follows the language prior, not the image. Report generation thus tells the same story as VQA and probing: ORCA preserves the clinically usable signal best, most visibly under heavy compression.

\subsection{Ablation study}
As a strong reference for how much a compressor could recover, we read out the full uncompressed token set with sinusoidal centroids (in Table~\ref{tab:ablation}). ORCA meets this reference to within a small margin, typically $0.01$ to $0.11$ $R^2$, all at $8$ to $64\times$ fewer tokens: it recovers most of the recoverable information at a fraction of the tokens. ORCA recovers this information in two ways: exposing spatial position, and aggregating tokens by content.

\paragraph{Position access.} Without explicit position, both Grid average and the aggregated tokens sit at the location floor, about $0.25$ on COLIPRI; the sinusoidal centroid lifts them to about $0.66$, close to uncompressed-with-centroid's $0.77$. The gap is not missing information but unreadable information: without explicit position, location survives only in the token ordering, which a readout cannot use~\cite{zaheer2017deepsets}. ORCA writes the region centroid into the token value, where the readout can use it, and closes most of that gap. The same position also lifts size, whose targets are spatially defined.

\begin{table}[!tb]
\centering\scriptsize
\setlength{\tabcolsep}{3pt}
\begin{tabular}{llccccc}
\toprule
encoder & variant & disease & size & density & location & texture \\
\midrule
\multicolumn{7}{c}{\textbf{CT-RATE}}\\
\midrule
\multirow{6}{*}{\shortstack[l]{COLIPRI\\ $B{=}216$}}
 & Grid average & 0.851 & 0.681 & 0.865 & 0.247 & 0.760 \\
 & \quad + centroid & 0.852 & 0.703 & 0.876 & 0.668 & 0.780 \\
\cmidrule(l){2-7}
 & aggregate & 0.852 & 0.683 & 0.907 & 0.251 & 0.784 \\
 & \quad + centroid & 0.853 & 0.711 & 0.908 & 0.657 & 0.786 \\
\rowcolor{gray!15} & \quad + organ (\ORCA{}) & 0.852 & 0.720 & 0.913 & 0.677 & 0.816 \\
\cmidrule(l){2-7}
 & Uncompressed + centroid & 0.849 & 0.791 & 0.930 & 0.773 & 0.829 \\
\midrule
\multirow{6}{*}{\shortstack[l]{CT-CLIP\\ $B{=}216$}}
 & Grid average & 0.743 & 0.612 & 0.570 & 0.529 & 0.566 \\
 & \quad + centroid & 0.758 & 0.665 & 0.601 & 0.663 & 0.582 \\
\cmidrule(l){2-7}
 & aggregate & 0.734 & 0.535 & 0.550 & 0.443 & 0.550 \\
 & \quad + centroid & 0.745 & 0.623 & 0.584 & 0.623 & 0.560 \\
\rowcolor{gray!15} & \quad + organ (\ORCA{}) & 0.748 & 0.649 & 0.586 & 0.645 & 0.564 \\
\cmidrule(l){2-7}
 & Uncompressed + centroid & 0.761 & 0.664 & 0.611 & 0.661 & 0.591 \\
\midrule
\multirow{6}{*}{\shortstack[l]{BTB3D\\ $B{=}512$}}
 & Grid average & 0.608 & 0.177 & 0.219 & 0.062 & 0.190 \\
 & \quad + centroid & 0.659 & 0.374 & 0.369 & 0.360 & 0.356 \\
\cmidrule(l){2-7}
 & aggregate & 0.613 & 0.159 & 0.177 & 0.034 & 0.153 \\
 & \quad + centroid & 0.676 & 0.332 & 0.320 & 0.238 & 0.336 \\
\rowcolor{gray!15} & \quad + organ (\ORCA{}) & 0.705 & 0.473 & 0.436 & 0.401 & 0.426 \\
\cmidrule(l){2-7}
 & Uncompressed + centroid & 0.686 & 0.465 & 0.480 & 0.448 & 0.460 \\
\midrule
\multicolumn{7}{c}{\textbf{Merlin}}\\
\midrule
\multirow{6}{*}{\shortstack[l]{SuPreM\\ $B{=}216$}}
 & Grid average & 0.787 & 0.462 & 0.681 & 0.395 & --- \\
 & \quad + centroid & 0.795 & 0.523 & 0.752 & 0.551 & --- \\
\cmidrule(l){2-7}
 & aggregate & 0.787 & 0.474 & 0.730 & 0.387 & --- \\
 & \quad + centroid & 0.798 & 0.550 & 0.785 & 0.584 & --- \\
\rowcolor{gray!15} & \quad + organ (\ORCA{}) & 0.795 & 0.544 & 0.795 & 0.600 & --- \\
\cmidrule(l){2-7}
 & Uncompressed + centroid & 0.806 & 0.622 & 0.845 & 0.709 & --- \\
\midrule
\multirow{6}{*}{\shortstack[l]{SegVol\\ $B{=}256$}}
 & Grid average & 0.746 & 0.466 & 0.626 & 0.378 & --- \\
 & \quad + centroid & 0.759 & 0.464 & 0.636 & 0.506 & --- \\
\cmidrule(l){2-7}
 & aggregate & 0.743 & 0.466 & 0.672 & 0.478 & --- \\
 & \quad + centroid & 0.760 & 0.499 & 0.700 & 0.553 & --- \\
\rowcolor{gray!15} & \quad + organ (\ORCA{}) & 0.761 & 0.588 & 0.688 & 0.596 & --- \\
\cmidrule(l){2-7}
 & Uncompressed + centroid & 0.780 & 0.675 & 0.790 & 0.668 & --- \\
\bottomrule
\end{tabular}
\caption{Component ablation. The last row, ``Uncompressed + centroid'', adds the position to the uncompressed tokens, where each centroid is a single grid patch. Per-cell standard deviations are ${\le}0.03$ on COLIPRI and SuPreM and up to $0.06$ on SegVol.}
\label{tab:ablation}
\end{table}

\paragraph{Anatomy access.} ORCA merges tokens by feature similarity instead of a fixed grid cell, with an organ mask steering the merge. The merge recovers content that Grid average blends away: on density, ORCA lifts its $0.865$ (COLIPRI) to $0.913$, close to uncompressed-with-centroid's $0.930$. The organ mask only helps -- adding it never meaningfully hurts (the largest drop is under $0.01$, within seed noise). It can usually be left on without per-encoder tuning.

\paragraph{Disease saturation.} The disease column moves differently from the others. Among the CT-RATE encoders, which share one abnormality target, ORCA lifts disease by only $+0.001$ (COLIPRI) and $+0.005$ (CT-CLIP) over Grid average, but by $+0.097$ on BTB3D. The two flat cases are the report-supervised encoders: their pretraining already read radiology reports, so the abnormality signal is redundantly encoded and the column is saturated -- no compressor adds to it. BTB3D, trained only to reconstruct, never saw that signal, leaving headroom that ORCA recovers. The Merlin encoders (SuPreM, SegVol), trained by segmentation without disease labels, show small positive gains on their own target, consistent with this reading.

The organ-guidance weight $\lambda$ and the position-encoding hyperparameters (frequency $F$, scale $s$) are robust knobs rather than tuned parameters: probe $R^2$ is nearly flat across four orders of magnitude of $\lambda$ and across the frequencies and scales we tried, with location the only responsive family. We fix $\lambda{=}0.5$ (COLIPRI), $\lambda{=}2$ (Merlin), and $F{=}4,s{=}2$ everywhere. The sweeps are in Appendix~\ref{app:position}.

\subsection{Efficiency}
The LLM computational gain depends mainly on the number of visual tokens that enter the LLM context, not on which compressor produced them. Table~\ref{tab:cost} therefore reports cost as a function of token budget. Prefill latency and KV-cache memory fall steeply, while peak memory falls only modestly but crosses the threshold that lets the budget run at all: on a commodity L4 (24\,GB) the uncompressed context otherwise does not fit. End-to-end report time moves little across budgets, since these short runs are decode dominated.

\begin{table}[!!htb]
\centering\scriptsize
\setlength{\tabcolsep}{4pt}
\begin{tabular}{lccccccc}
\toprule
 & & & & \multicolumn{2}{c}{prefill (ms)} & \multicolumn{2}{c}{s / report} \\
\cmidrule(lr){5-6}\cmidrule(lr){7-8}
$B$ & FLOPs (T) & KV (MB) & peak (GB) & B200 & L4 & B200 & L4 \\
\midrule
13{,}824 & 324 & 1820 & 27.1 & 371.5 & OOM & 1.84 & OOM \\
216 & 4.5 & 36.7 & 16.3 & 11.8 & 132.9 & 1.43 & 8.06 \\
64 & 2.1 & 16.8 & 16.2 & 11.5 & \phantom{0}80.5 & 1.43 & 7.98 \\
27 & 1.5 & 11.9 & 16.1 & 11.6 & \phantom{0}78.3 & 1.43 & 7.98 \\
8 & 1.2 & \phantom{0}9.4 & 16.1 & 11.6 & \phantom{0}77.0 & 1.43 & 7.98 \\
\bottomrule
\end{tabular}
\caption{LLM cost vs.\ token budget $B$ on Llama-3.1-8B (bf16), for a B200 and a commodity L4 (24\,GB). Each latency is the median of three alternating sweeps (warmup discarded); FLOPs, KV-cache, and peak memory depend on $B$ only, not the GPU.}
\label{tab:cost}
\end{table}

\section{Discussion and Limitations}
\paragraph{Preserve the anatomical information.}
For a 3D CT embedding, the goal of compression is to preserve the information a downstream model needs, not to reconstruct the original tokens as faithfully as possible. That information is spatial and heterogeneous. Because it is anatomical rather than task-specific, even a training-free operator preserves most of it. By writing each region's centroid and merging by content, ORCA stays within a small margin of the uncompressed tokens at matched budgets. We test only two task types here, but the approach should carry over to other 3D CT tasks.

\paragraph{Attributes beyond disease.}
The dominant way to score a 3D CT model is report generation and the VQA derived from it, with its metrics anchored on disease findings. This does not fully cover the attributes a CT volume carries. To reach the rest, we evaluate on the mask-derived measurement VQA benchmark of \citet{liang2026cheap}, whose labels come from anatomical measurements, are defined for every scan, and span size, density, and location alongside disease. Across encoders, most are optimized and benchmarked mainly for disease, leaving these other attributes largely unmeasured.

\paragraph{Keeping the budget adjustable.}
A natural alternative is to pool tokens directly within each organ mask, but the obstacle is budget, not accuracy. Hard organ pooling emits one token per organ or organ slice, so the token count is fixed by anatomy rather than chosen by the user. ORCA instead makes the budget a free parameter, and the more tokens it is given, the more information it preserves.


\section{Conclusion}
We presented ORCA, a training-free compressor that aggregates 3D CT visual tokens into connected, content-adaptive regions and writes each region's position back into the token. Across two datasets and five encoders it improves over Grid average and matched token-reduction baselines, and stays within a small margin of the uncompressed tokens at $8$ to $64\times$ fewer tokens, with no compression network trained. The principle is simple: what compression preserves is governed by what the downstream reader can read, so a good compressor makes position explicit and aggregates by content rather than by a fixed grid. Structure-preserving aggregation with position re-injection is a strong drop-in replacement for Grid average in 3D CT vision-language pipelines.
\bibliography{refs}

\appendix
\setcounter{secnumdepth}{2}
\color{black}
\twocolumn[{%
\begin{center}
{\LARGE\bfseries Supplementary Material}\\[0.45em]
{\large\bfseries ORCA: ORgan-Centroid Aggregation for Training-Free 3D CT Visual Token Compression}
\end{center}
\vspace{0.6em}
}]

\section{Experimental setup}

\subsection{Datasets and encoders}

\begin{table}[!htb]\centering\scriptsize
\setlength{\tabcolsep}{5pt}
\begin{tblr}{
    colspec = {lcc}, colsep = 6pt, rowsep = 1.5pt,
    row{2-5} = {gray!12},
    hline{1,10} = {0.08em},
    hline{2}    = {0.05em},
    hline{6}    = {0.05em},
  }
   & CT-RATE & Merlin \\
  anatomy & chest & abdomen \\
  contrast & non-contrast & portal-venous \\
  patients & 21{,}304 & 18{,}317 \\
  abnormality labels & 18 & 30 \\
  CT volumes & 25{,}692 & 25{,}494 \\
  \quad {train / valid} & 24{,}128 / 1{,}564 & 15{,}309 / 5{,}055 \\
  reconstructions & 50{,}188 & 25{,}494 \\
  \quad {train / valid} & 47{,}149 / 3{,}039 & 15{,}309 / 5{,}055 \\
  \end{tblr}
\caption{Dataset scale and imaging characteristics of CT-RATE~\cite{hamamci2024ctrate} and Merlin~\cite{blankemeier2026merlin}. Each CT-RATE volume may correspond to multiple reconstructions generated under different reconstruction settings, whereas each Merlin scan corresponds to a single reconstructed volume.}
\label{tab:app_datasets}
\end{table}

The dataset scale and splits are reported in Table~\ref{tab:app_datasets}. Each scan in both datasets is paired with a radiology report. Abnormality labels come from RadBERT~\cite{radbert} and organ masks from TotalSegmentator~\cite{totalsegmentator}. CT-CLIP and BTB3D train on the reconstruction-level split, whereas COLIPRI uses the volume-level split; Merlin uses a $20{,}364$-scan subset.

\begin{figure*}[!t]
\centering
\includegraphics[width=\textwidth]{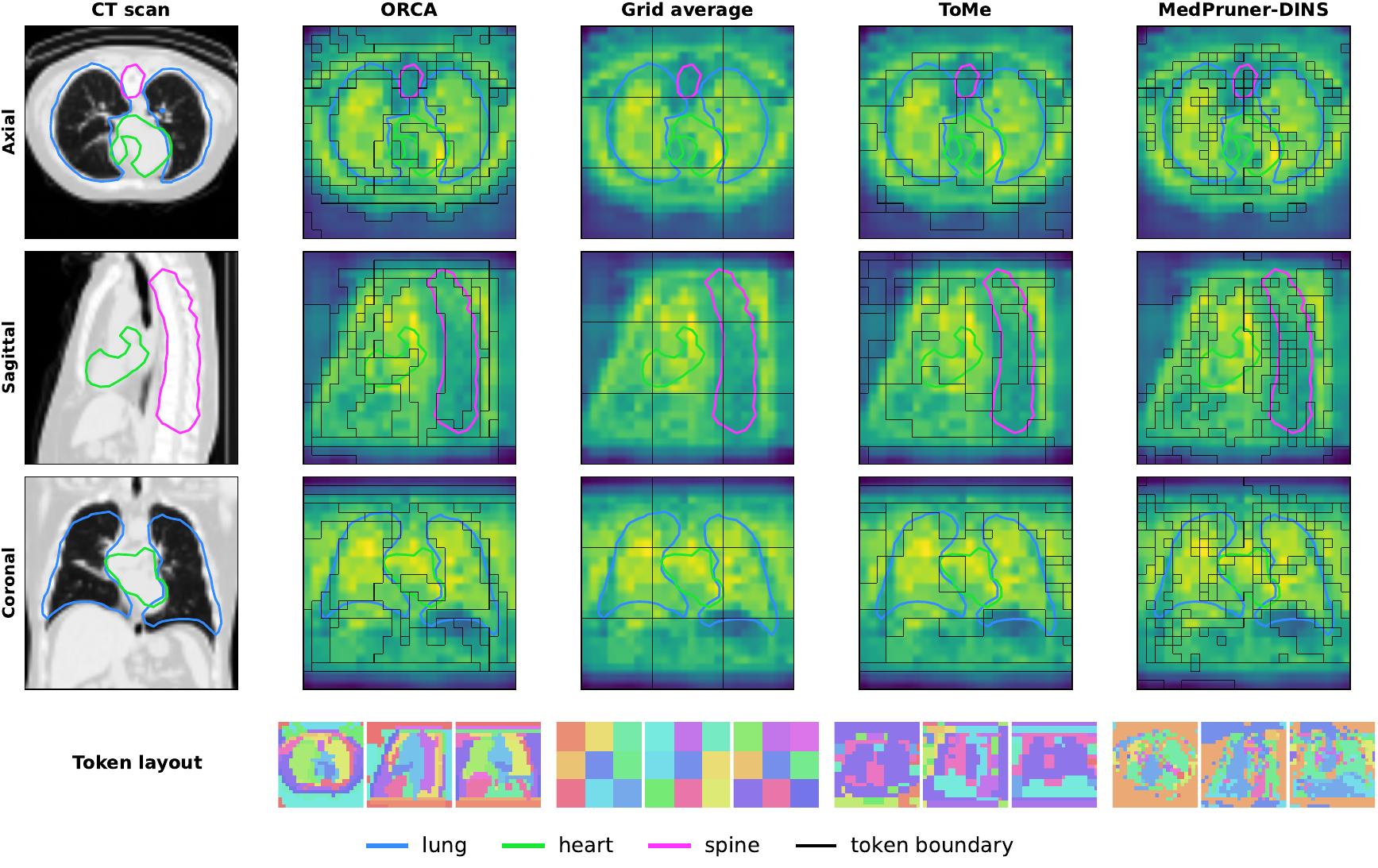}
\caption{Case study on one CT volume (COLIPRI, $B{=}27$) in three anatomical views. The first column is the CT scan with lung, heart, and spine contours; the remaining columns overlay each compressor's token boundaries on the embedding-norm heatmap, and the bottom row recolors the same partitions so that each token receives one color. ORCA aggregates tokens into spatially contiguous regions that follow the organ contours.}
\label{fig:case_study}
\end{figure*}

\subsection{Measurement VQA benchmark}
\label{app:vqabench}

Our probing and visual question answering (VQA) experiments use the measurement VQA benchmark introduced by \citet{liang2026cheap}. The benchmark comprises a set of image-derived anatomical attributes whose reference values are computed directly from each CT volume, its corresponding segmentation masks, and Hounsfield unit (HU) values. Because the labels are generated from deterministic image measurements, they are reproducible and avoid noise introduced by variations in radiology report wording or LLM generation.

The benchmark is designed to evaluate token compression according to the anatomical information retained from the CT image. It covers attributes such as size, density, location, and texture, which are often not explicitly quantified in radiology reports. Each question targets a single organ, allowing VQA accuracy to reflect how well the compressed representation preserves localized anatomical details. The same attributes are used as regression or AUROC evaluation targets in the probing experiments and are converted into multiple-choice questions based on population tertiles or predefined clinical thresholds for VQA. The complete benchmark construction, statistical characterization, and validation are provided by \citet{liang2026cheap}. Table~\ref{tab:app_attr_detail} details every attribute, and Figure~\ref{fig:vqa_q_examples} gives representative question--answer examples.

\begin{tcolorbox}[enhanced, breakable, width=0.99\columnwidth, colback=gray!5, colframe=gray!45,
  boxrule=0.5pt, arc=3pt, left=8pt, right=8pt, top=7pt, bottom=7pt, fontupper=\small]
\textbf{Q1.}~``How would you characterize the cardiothoracic (heart-to-lung) size ratio?''\\[2pt]
\hspace*{1.2em}(A) High\\
\hspace*{1.2em}(B) Moderate\\
\hspace*{1.2em}\kern-\fboxsep\colorbox{green!15}{\textbf{(C) Low}}

\medskip
\textbf{Q2.}~``On this scan, the aortic diameter appears:''\\[2pt]
\hspace*{1.2em}\kern-\fboxsep\colorbox{green!15}{\textbf{(A) Normal caliber, $<40$~mm}}\\
\hspace*{1.2em}(B) Dilated, $\geq 40$~mm

\medskip
\textbf{Q3.}~``The aortic wall calcification burden on this CT is best described as:''\\[2pt]
\hspace*{1.2em}\kern-\fboxsep\colorbox{green!15}{\textbf{(A) Present}}\\
\hspace*{1.2em}(B) Absent

\medskip
\textbf{Q4.}~``On this scan, the heart horizontal (left-right) position appears:''\\[2pt]
\hspace*{1.2em}(A) High\\
\hspace*{1.2em}(B) Low\\
\hspace*{1.2em}\kern-\fboxsep\colorbox{green!15}{\textbf{(C) Moderate}}

\medskip
\textbf{Q5.}~``On this scan, the lung 15th-percentile density (emphysema index) appears:''\\[2pt]
\hspace*{1.2em}(A) No significant emphysema, $\geq -950$~HU\\
\hspace*{1.2em}\kern-\fboxsep\colorbox{green!15}{\textbf{(B) Emphysema present, $<-950$~HU}}
\end{tcolorbox}
\captionof{figure}{Representative measurement VQA questions and reference answers. The reference answer is highlighted in green.}
\label{fig:vqa_q_examples}

\subsection{Implementation details}
\label{app:impl}
\paragraph{Probes.} Each family is read out from the frozen compressed tokens by a lightweight attention-pooling head~\cite{ilse2018attention} followed by a small MLP, trained with AdamW~\cite{loshchilov2019decoupled}; every probing number is the mean over three seeds. Two comparisons instead use a higher-capacity head so that no method is bottlenecked by the probe: the uncompressed reference rows in Table~\ref{tab:ablation}, and the Merlin centroid-encoding comparison (Table~\ref{tab:centbaselines_merlin}). Encoder features are normalized to the encoder's scale before the readout. Exact hyperparameters are in the released code.

\paragraph{Text generation.} The generation model is a LLaVA-style pipeline~\cite{liu2023visual} with a Llama-3.1-8B-Instruct backbone~\cite{llama3herd} and a two-layer projector. Training is two-stage: a projector warm-up with the backbone frozen, then LoRA fine-tuning~\cite{hu2021lora} of the backbone. Each encoder--compressor combination is trained separately. Validation is a full-set generation pass each epoch and the reported number is the best epoch.
\paragraph{Hardware.} We train the VQA and text-generation models on a single NVIDIA B200 GPU (180\,GB) with bf16 mixed precision. All other experiments, including probing and preprocessing, run on a single NVIDIA L4 GPU (24\,GB). ORCA compression itself is CPU-only and needs no GPU, so the only heavy hardware requirement is the language-model fine-tuning.

\begin{table*}[!htb]\centering\scriptsize
\begin{tblr}{
  colspec = {l cccccccc},
  colsep = 5pt,
  rowsep = 1.1pt,
  cell{2}{2} = {c=4}{c},
  cell{2}{6} = {c=4}{c},
  cell{12}{2} = {c=4}{c},
  cell{12}{6} = {c=4}{c},
  vline{6} = {0.03em},
  row{11} = {gray!15},
  row{19} = {gray!15},
  hline{1} = {0.08em},
  hline{2} = {2-9}{0.03em},
  hline{3} = {0.05em},
  hline{12} = {0.06em},
  hline{13} = {0.05em},
  hline{20} = {0.08em},
}
 & size & density & location & texture & size & density & location & texture \\
COLIPRI & $B{=}27$ &  &  &  & $B{=}216$ &  &  &  \\
Grid average & \FIN{0.676}{0.005} & \FIN{0.806}{0.005} & \FIN{0.266}{0.023} & \FIN{0.683}{0.008} & \FIN{0.681}{0.002} & \FIN{0.865}{0.004} & \FIN{0.247}{0.014} & \FIN{0.760}{0.025} \\
\quad + centroid & \FIN{0.685}{0.010} & \FIN{0.815}{0.006} & \FIN{0.489}{0.020} & \FIN{0.705}{0.017} & \FIN{0.703}{0.011} & \FIN{0.876}{0.010} & \FIN{0.668}{0.030} & \FIN{0.780}{0.010} \\
DivPrune & \FIN{0.652}{0.005} & \FIN{0.854}{0.004} & \FIN{0.228}{0.002} & \FIN{0.741}{0.010} & \FIN{0.680}{0.005} & \FIN{0.887}{0.002} & \FIN{0.260}{0.009} & \FIN{0.753}{0.015} \\
\quad + centroid & \FIN{0.666}{0.005} & \FIN{0.852}{0.002} & \FIN{0.510}{0.017} & \FIN{0.764}{0.013} & \FIN{0.709}{0.006} & \FIN{0.894}{0.002} & \FIN{0.674}{0.001} & \FIN{0.794}{0.008} \\
MedPruner-DINS & \FIN{0.635}{0.003} & \FIN{0.848}{0.005} & \FIN{0.175}{0.004} & \FIN{0.706}{0.027} & \FIN{0.673}{0.007} & \FIN{0.903}{0.003} & \FIN{0.233}{0.017} & \FIN{0.765}{0.022} \\
\quad + centroid & \FIN{0.645}{0.001} & \FIN{0.854}{0.005} & \FIN{0.393}{0.029} & \FIN{0.737}{0.018} & \FIN{0.703}{0.008} & \FIN{0.906}{0.003} & \FIN{0.519}{0.012} & \FIN{0.792}{0.015} \\
ToMe & \FIN{0.617}{0.002} & \FIN{0.786}{0.011} & \FIN{0.172}{0.004} & \FIN{0.647}{0.001} & \FIN{0.636}{0.010} & \FIN{0.845}{0.004} & \FIN{0.220}{0.022} & \FIN{0.714}{0.019} \\
\quad + centroid & \FIN{0.618}{0.003} & \FIN{0.792}{0.003} & \FIN{0.212}{0.026} & \FIN{0.659}{0.008} & \FIN{0.651}{0.002} & \FIN{0.846}{0.002} & \FIN{0.338}{0.019} & \FIN{0.743}{0.017} \\
\ORCA{} & \FIN{\textbf{0.691}}{0.011} & \FIN{\textbf{0.895}}{0.004} & \FIN{\textbf{0.622}}{0.005} & \FIN{\textbf{0.775}}{0.009} & \FIN{\textbf{0.720}}{0.005} & \FIN{\textbf{0.913}}{0.002} & \FIN{\textbf{0.677}}{0.013} & \FIN{\textbf{0.816}}{0.001} \\
BTB3D & $B{=}27$ &  &  &  & $B{=}216$ &  &  &  \\
Grid average & \FIN{0.183}{0.014} & \FIN{0.182}{0.008} & \FIN{0.070}{0.004} & \FIN{0.194}{0.005} & \FIN{0.188}{0.013} & \FIN{0.205}{0.018} & \FIN{0.071}{0.003} & \FIN{0.185}{0.008} \\
\quad + centroid & \FIN{0.324}{0.005} & \FIN{0.302}{0.016} & \FIN{0.183}{0.008} & \FIN{0.316}{0.011} & \FIN{0.378}{0.008} & \FIN{0.372}{0.005} & \FIN{0.341}{0.007} & \FIN{0.364}{0.003} \\
DivPrune & \FIN{0.149}{0.001} & \FIN{0.181}{0.001} & \FIN{0.056}{0.003} & \FIN{0.164}{0.003} & \FIN{0.181}{0.007} & \FIN{0.223}{0.010} & \FIN{0.069}{0.004} & \FIN{0.201}{0.009} \\
\quad + centroid & \FIN{0.256}{0.001} & \FIN{0.276}{0.000} & \FIN{0.190}{0.002} & \FIN{0.266}{0.019} & \FIN{0.349}{0.005} & \FIN{0.368}{0.014} & \FIN{0.309}{0.011} & \FIN{0.370}{0.010} \\
ToMe & \FIN{0.100}{0.005} & \FIN{0.123}{0.002} & \FIN{0.039}{0.001} & \FIN{0.119}{0.003} & \FIN{0.126}{0.003} & \FIN{0.157}{0.003} & \FIN{0.046}{0.002} & \FIN{0.144}{0.004} \\
\quad + centroid & \FIN{0.114}{0.003} & \FIN{0.124}{0.002} & \FIN{0.041}{0.001} & \FIN{0.161}{0.005} & \FIN{0.183}{0.006} & \FIN{0.196}{0.007} & \FIN{0.079}{0.003} & \FIN{0.236}{0.011} \\
\ORCA{} & \FIN{\textbf{0.420}}{0.002} & \FIN{\textbf{0.345}}{0.013} & \FIN{\textbf{0.564}}{0.009} & \FIN{\textbf{0.361}}{0.003} & \FIN{\textbf{0.441}}{0.004} & \FIN{\textbf{0.405}}{0.004} & \FIN{\textbf{0.434}}{0.025} & \FIN{\textbf{0.402}}{0.006} \\
\end{tblr}
\caption{Centroid encoding on CT-RATE probing. Each baseline is shown without and with the sinusoidal centroid encoding ORCA uses; DINS does not apply to BTB3D, whose reconstruction features carry no attention saliency. The best value in each column, within each encoder, is in bold.}
\label{tab:centbaselines}
\end{table*}

\begin{table}[!htb]\centering\scriptsize
\begin{tblr}{
  colspec = {lcccc},
  colsep  = 4pt,
  rowsep  = 1.1pt,
  cell{2}{1} = {c=5}{c},
  cell{6}{1} = {c=5}{c},
  row{5}     = {gray!15},
  row{9}     = {gray!15},
  hline{1,10} = {0.08em},
  hline{2}   = {0.05em},
  hline{6}   = {0.03em},
}
method & size & density & location & texture \\
$B{=}27$ & & & & \\
Grid average & 0.759 & 0.768 & 0.601 & 0.783 \\
\quad + centroid & 0.756 & 0.781 & 0.615 & 0.788 \\
\ORCA{} & 0.760 & 0.838 & 0.675 & 0.829 \\
$B{=}216$ & & & & \\
Grid average & 0.766 & 0.807 & 0.649 & 0.795 \\
\quad + centroid & 0.763 & 0.811 & 0.701 & 0.778 \\
\ORCA{} & 0.778 & 0.847 & 0.721 & 0.835 \\
\end{tblr}
\caption{Downstream measurement VQA accuracy on CT-RATE/COLIPRI.}
\label{tab:centgen_vqa}
\end{table}
\subsection{Baseline adaptations}
\label{app:baseline_adapt}
\color{black}
All baselines are placed at the same encoder-output interface as ORCA. They receive the raw 3D token grid and produce a shorter token sequence before the projector. Methods differ in whether they expose an adjustable target count. Grid average, DivPrune, ToMe, and MedPruner-DINS do, and we force each to emit the same $B$ tokens as ORCA. Slice pooling and MedRegion-CT pooling do not: their counts are fixed by construction, by encoder depth and anatomy respectively, so we report each at its natural operating point. Several of these methods were designed for 2D slice-stack encoders or include learned components in their original form. We adapt each to our 3D-native, training-free interface while keeping its core principle.

\paragraph{Grid average.} Grid average pools regular cells of the encoder grid and replaces each cell by the mean of its visual embeddings. For grids and budgets that admit an integer stride, this is non-overlapping $r{\times}r{\times}r$ block averaging. For non-cubic grids or budgets that do not correspond to an integer stride, we use 3D adaptive average pooling to the nearest aspect-preserving grid. The output dimension stays equal to the encoder feature dimension.

\paragraph{Slice pooling.} Slice pooling averages each axial token plane into one token. For a grid of shape $T{\times}H{\times}W$, it returns $T$ tokens, each the mean over one $H{\times}W$ plane. It is a coarse slice-level baseline with no within-plane or organ structure. Its token count is fixed by the encoder depth, so we report it at its actual count.

\paragraph{DivPrune.} DivPrune~\cite{alvar2025divprune} is adapted as a feature-diversity selector on the 3D encoder tokens. We first apply uniform 3D average pooling to a $1{,}024$-token candidate set, then select $B$ tokens by farthest-first diversity in cosine-normalized feature space. This keeps the diversity principle of DivPrune. The method was designed to prune the tokens of a 2D-slice VLM, and we adapt it to operate on the dense 3D grid. It is a pure selection baseline: tokens not selected do not contribute to the output.

\paragraph{ToMe.} ToMe~\cite{bolya2023tome} is adapted as feature-similarity merging on the flattened 3D token sequence. We first uniformly pool the dense grid to a $1{,}024$-token candidate set, then run bipartite cosine matching and repeatedly merge the most similar token pairs until $B$ tokens remain. This baseline merges rather than drops tokens.

\paragraph{MedRegion-CT pooling.} MedRegion-CT's original system learns a SlowFast tokenizer with pseudo-mask guidance and structured prompts. Only its region-pooling recipe transfers to our fixed-encoder, training-free interface, so we reproduce that part faithfully~\cite{kyung2025medregionct}: one global token per axial plane plus one region token for each organ present in that plane. The global token is the mean over all tokens in the plane. The region token is an organ-occupancy-weighted mean over tokens in the same plane. We do not force this method to match a target budget, because its token count is determined by how many slice--organ pairs are present. It therefore emits a variable count, $\bar N{=}549$ on CT-RATE, which we report at that single operating point.

\paragraph{MedPruner-DINS.} MedPruner~\cite{liu2026medpruner} has two stages: inter-slice filtering and token-level Dynamic Information Nucleus Selection (DINS). Its inter-slice filtering stage assumes a 2D slice-stack encoder, where tokens are produced separately per slice. Our encoders are 3D-native and already fold the depth axis into one volumetric token grid, so we drop the inter-slice filtering and keep only the DINS stage. We score tokens by the attention they receive in the vision encoder, keep the top $B$ tokens, and fold the remaining tokens into their most similar kept token by cosine similarity before averaging. This produces exactly $B$ output tokens while preserving MedPruner's idea that low-attention tokens can still contribute through residual merging. We run this baseline only where the encoder exposes a clean global per-token attention score. This excludes SuPreM, whose Swin-UNETR backbone uses windowed local attention.

\section{Additional results and analysis}

\subsection{Case study}
\label{app:casestudy}
Figure~\ref{fig:case_study} illustrates, on a single volume, how each compressor lays its tokens over the anatomy. ORCA aggregates tokens into spatially contiguous regions that follow the organ contours. Grid average instead imposes a fixed grid that cuts across organ borders, and ToMe and MedPruner-DINS merge tokens by feature similarity with no spatial regularity, scattering each token into many disconnected fragments. The bottom-row token-layout maps make this contrast clear: ORCA keeps each token to a single connected region aligned with anatomy, an advantage the pooling and pruning baselines lack.

\begin{figure}[!tb]
\centering
\includegraphics[width=\columnwidth]{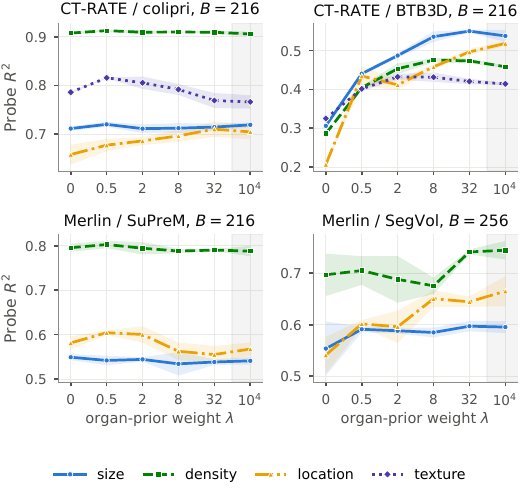}
\caption{Sensitivity of probe $R^2$ to the organ-guidance weight $\lambda$.}
\label{fig:lambda_sweep}
\end{figure}

\subsection{Centroid encoding}
\label{app:centgen}
To separate the contribution of explicit position encoding from that of token aggregation, we give every baseline the same sinusoidal centroid encoding ORCA uses, applied to each method's own token centroids, and re-evaluate probing (Tables~\ref{tab:centbaselines} and~\ref{tab:centbaselines_merlin}) and downstream VQA (Table~\ref{tab:centgen_vqa}). Because a centroid can be computed for any retained or merged token, this equalizes position across methods, so any remaining gap reflects how the tokens are formed rather than whether they carry position. On Merlin, whose encoders are sensitive to probe capacity, we read out with a high-capacity probe so that no baseline is limited by the probe; COLIPRI is capacity-insensitive and uses the default probe.

Adding position helps every baseline, and the gain concentrates almost entirely on location. On COLIPRI it lifts Grid average from $0.247$ to $0.668$ and DivPrune from $0.260$ to $0.674$ at $B{=}216$, while the content families move by only a few points. Position injection is thus a general, method-agnostic benefit for the one family that depends on it. ToMe is the exception, with a far smaller location gain: because it scatters each token across the volume, its centroid is unrepresentative and the encoding it receives is close to noise, foreshadowing that a centroid is only useful when the token it summarizes is spatially coherent. Even with position equalized, ORCA still leads, most clearly on the content families and under strong compression. On COLIPRI it keeps a margin on size, density, and texture at both budgets, for example density $0.913$ against $0.906$ for the best centroid-augmented baseline, and its location advantage is large at $B{=}27$ ($0.622$ against $0.510$) though it narrows to a near tie at $B{=}216$, where position alone nearly suffices. The gap that the centroid cannot close is therefore on content, and it comes from ORCA's adaptive, organ-aligned aggregation rather than from position. This is starkest on the weak reconstruction encoder BTB3D, where centroid-augmented baselines still recover little location signal ($\sim\!0.18$) while ORCA reaches $0.564$; on Merlin, ORCA leads every column.

\begin{figure}[!tb]
\centering
\includegraphics[width=\linewidth]{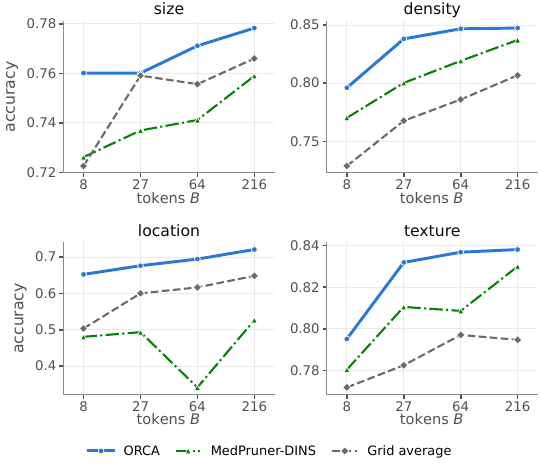}
\caption{VQA accuracy across token budgets on CT-RATE with the COLIPRI encoder.}
\label{fig:vqa_budget}
\end{figure}

\begin{table*}[!htb]\centering\scriptsize
\begin{tblr}{
  colspec = {l cccccc},
  colsep = 5pt,
  rowsep = 1.1pt,
  cell{2}{2}  = {c=3}{c},
  cell{2}{5}  = {c=3}{c},
  cell{10}{2} = {c=3}{c},
  cell{10}{5} = {c=3}{c},
  vline{5} = {0.03em},
  row{9} = {gray!15},
  row{17} = {gray!15},
  hline{1} = {0.08em},
  hline{2} = {2-7}{0.03em},
  hline{3} = {0.05em},
  hline{10} = {0.06em},
  hline{11} = {0.05em},
  hline{18} = {0.08em},
}
 & size & density & location & size & density & location \\
SuPreM & $B{=}64$ &  &  & $B{=}216$ &  &  \\
Grid average & \FIN{0.519}{0.003} & \FIN{0.741}{0.013} & \FIN{0.537}{0.015} & \FIN{0.544}{0.009} & \FIN{0.786}{0.006} & \FIN{0.551}{0.018} \\
\quad + centroid & \FIN{0.531}{0.003} & \FIN{0.736}{0.011} & \FIN{0.566}{0.005} & \FIN{0.578}{0.008} & \FIN{0.795}{0.004} & \FIN{0.655}{0.015} \\
DivPrune & \FIN{0.521}{0.006} & \FIN{0.770}{0.006} & \FIN{0.501}{0.011} & \FIN{0.569}{0.010} & \FIN{0.803}{0.018} & \FIN{0.577}{0.021} \\
\quad + centroid & \FIN{0.528}{0.008} & \FIN{0.752}{0.005} & \FIN{0.577}{0.000} & \FIN{0.589}{0.005} & \FIN{0.805}{0.006} & \FIN{0.665}{0.010} \\
ToMe & \FIN{0.350}{0.017} & \FIN{0.624}{0.002} & \FIN{0.261}{0.002} & \FIN{0.439}{0.019} & \FIN{0.705}{0.008} & \FIN{0.351}{0.013} \\
\quad + centroid & \FIN{0.348}{0.010} & \FIN{0.619}{0.008} & \FIN{0.263}{0.012} & \FIN{0.449}{0.006} & \FIN{0.716}{0.004} & \FIN{0.380}{0.010} \\
\ORCA{} & \FIN{\textbf{0.585}}{0.010} & \FIN{\textbf{0.836}}{0.003} & \FIN{\textbf{0.700}}{0.007} & \FIN{\textbf{0.609}}{0.004} & \FIN{\textbf{0.851}}{0.005} & \FIN{\textbf{0.721}}{0.011} \\
SegVol & $B{=}32$ &  &  & $B{=}256$ &  &  \\
Grid average & \FIN{0.599}{0.007} & \FIN{0.722}{0.004} & \FIN{0.549}{0.021} & \FIN{0.662}{0.011} & \FIN{0.778}{0.009} & \FIN{0.662}{0.008} \\
\quad + centroid & \FIN{0.585}{0.012} & \FIN{0.720}{0.005} & \FIN{0.524}{0.007} & \FIN{0.633}{0.006} & \FIN{0.774}{0.003} & \FIN{0.620}{0.011} \\
DivPrune & \FIN{0.562}{0.011} & \FIN{0.706}{0.009} & \FIN{0.552}{0.009} & \FIN{0.677}{0.005} & \FIN{0.785}{0.020} & \FIN{0.700}{0.009} \\
\quad + centroid & \FIN{0.551}{0.006} & \FIN{0.716}{0.002} & \FIN{0.531}{0.033} & \FIN{0.659}{0.006} & \FIN{0.806}{0.014} & \FIN{0.675}{0.053} \\
ToMe & \FIN{0.506}{0.005} & \FIN{0.651}{0.005} & \FIN{0.468}{0.012} & \FIN{0.633}{0.010} & \FIN{0.749}{0.007} & \FIN{0.683}{0.005} \\
\quad + centroid & \FIN{0.478}{0.011} & \FIN{0.638}{0.005} & \FIN{0.466}{0.006} & \FIN{0.620}{0.006} & \FIN{0.754}{0.002} & \FIN{0.678}{0.025} \\
\ORCA{} & \FIN{\textbf{0.680}}{0.001} & \FIN{\textbf{0.818}}{0.010} & \FIN{\textbf{0.710}}{0.016} & \FIN{\textbf{0.685}}{0.004} & \FIN{\textbf{0.833}}{0.010} & \FIN{\textbf{0.747}}{0.007} \\
\end{tblr}
\caption{Centroid encoding on Merlin probing, read out with a high-capacity probe (dimension $1024$) so that no method is limited by the probe. DINS does not apply to SuPreM's windowed attention, and Merlin has no texture family. The best value in each column, within each encoder, is in bold.}
\label{tab:centbaselines_merlin}
\end{table*}

\subsection{Sensitivity to the organ mask}
\label{app:position}
Figure~\ref{fig:lambda_sweep} sweeps the organ-guidance weight $\lambda$ from $0$, no mask, to $10^4$, the organ-dominated limit that reduces to hard within-organ pooling, for each family on all four encoders. Across a wide range the curves vary only gradually with no sharp optimum, and even the organ-dominated limit does not collapse them, so the mask weight is a robust knob rather than a fragile tuned parameter. The family that consistently responds is location, which rises with $\lambda$ on every encoder, for instance $0.66$ to $0.71$ on COLIPRI and $0.54$ to $0.67$ on SegVol, consistent with the mask supplying organ identity that most directly aids localization. The gains are uneven across encoders: the COLIPRI curves move only a little while the BTB3D curves rise more. We observe and speculate that COLIPRI's embeddings are already distributed much like the organs themselves, so the organ mask adds little, whereas BTB3D's embeddings look closer to random with no clear organ structure, so the mask helps more.

\subsection{Budget scaling}
\label{app:vqabudget}
Figure~\ref{fig:vqa_budget} plots best epoch VQA accuracy against the token budget for the three compressors on CT-RATE with the COLIPRI encoder, one panel per family. ORCA leads at every budget in every family, and its margin is widest on location and density while size and texture are close to saturated. The scaling is efficient: ORCA at $B{=}8$ already matches Grid average at $B{=}216$ across all four families to within $0.01$ accuracy, for instance $0.653$ against $0.649$ on location, reaching the same downstream accuracy with $27{\times}$ fewer tokens. The baselines are less stable at the extremes, and MedPruner-DINS drops to near chance on location at $B{=}64$.

\begin{figure}[!htb]
\centering
\includegraphics[width=\columnwidth]{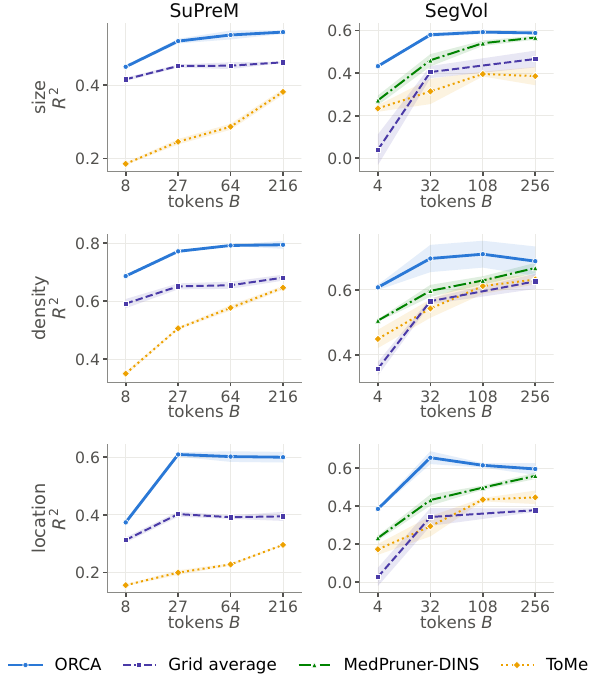}
\caption{Probing budget curves on the Merlin encoders, SuPreM and SegVol.}
\label{fig:budget_merlin}
\end{figure}

Figure~\ref{fig:budget_merlin} repeats the analysis with intrinsic probing $R^2$ on the two Merlin encoders, SuPreM and SegVol, over the size, density, and location families. The same effects appear on a second encoder family under a different readout: ORCA leads at nearly every budget, its advantage is again largest on location, and it saturates early. On SuPreM, ORCA at $B{=}8$ already exceeds Grid average at $B{=}216$ on density, $0.687$ against $0.681$, and its location score at $B{=}27$ of $0.610$ far exceeds Grid average at $B{=}216$ of $0.395$. Grid average is the weakest at very small budgets, collapsing to $R^2$ near zero on SegVol at $B{=}4$.

\subsection{Held-out test generalization}
\label{app:heldout}
The probe readouts are trained on data, so we check that the method ranking is not specific to the official validation split. We reserve $2{,}000$ volumes from the CT-RATE training pool as an independent held-out test set, disjoint from both the probe-training volumes and the validation split, and re-evaluate the CT-RATE/COLIPRI comparison there. Table~\ref{tab:heldout} reports the held-out scores together with their gap to validation. The ranking is preserved: ORCA leads every family at both budgets, most strikingly on location where it reaches $0.57$ to $0.65$ against $0.13$ to $0.25$ for the baselines, while disease is saturated and ties. The held-out and validation scores differ only slightly and unsystematically, at most about $0.05$ and usually under $0.02$, so the probes are not overfit to the validation split and ORCA's advantage generalizes.

\begin{table*}[!htb]
\centering\scriptsize
\setlength{\tabcolsep}{4.2pt}
\begin{tabular}{llccccc}
\toprule
method & $B$ & disease & size & density & location & texture \\
\midrule
\multirow{2}{*}{Grid average}
 & 27 & \FIN{0.848}{0.001}\,{\tiny\textcolor{gray}{$-$0.002}} & \FIN{0.664}{0.006}\,{\tiny\textcolor{gray}{$-$0.011}} & \FIN{0.797}{0.009}\,{\tiny\textcolor{gray}{$-$0.000}} & \FIN{0.229}{0.032}\,{\tiny\textcolor{gray}{$-$0.009}} & \FIN{0.682}{0.024}\,{\tiny\textcolor{gray}{+0.010}} \\
 & 216 & \FIN{0.849}{0.001}\,{\tiny\textcolor{gray}{$-$0.002}} & \FIN{0.676}{0.011}\,{\tiny\textcolor{gray}{$-$0.008}} & \FIN{0.858}{0.009}\,{\tiny\textcolor{gray}{+0.001}} & \FIN{0.246}{0.008}\,{\tiny\textcolor{gray}{$-$0.015}} & \FIN{0.707}{0.021}\,{\tiny\textcolor{gray}{$-$0.018}} \\
\midrule
\multirow{2}{*}{MedPruner-DINS~\cite{liu2026medpruner}}
 & 27 & \FIN{0.844}{0.001}\,{\tiny\textcolor{gray}{$-$0.004}} & \FIN{0.625}{0.006}\,{\tiny\textcolor{gray}{$-$0.008}} & \FIN{0.844}{0.004}\,{\tiny\textcolor{gray}{+0.002}} & \FIN{0.144}{0.005}\,{\tiny\textcolor{gray}{$-$0.025}} & \FIN{0.679}{0.014}\,{\tiny\textcolor{gray}{$-$0.015}} \\
 & 216 & \FIN{0.848}{0.001}\,{\tiny\textcolor{gray}{$-$0.002}} & \FIN{0.670}{0.003}\,{\tiny\textcolor{gray}{$-$0.009}} & \FIN{0.900}{0.001}\,{\tiny\textcolor{gray}{$-$0.001}} & \FIN{0.199}{0.007}\,{\tiny\textcolor{gray}{$-$0.031}} & \FIN{0.729}{0.012}\,{\tiny\textcolor{gray}{$-$0.025}} \\
\midrule
\multirow{2}{*}{ToMe~\cite{bolya2023tome}}
 & 27 & \FIN{0.842}{0.001}\,{\tiny\textcolor{gray}{$-$0.002}} & \FIN{0.601}{0.004}\,{\tiny\textcolor{gray}{$-$0.012}} & \FIN{0.783}{0.012}\,{\tiny\textcolor{gray}{$-$0.001}} & \FIN{0.134}{0.004}\,{\tiny\textcolor{gray}{$-$0.037}} & \FIN{0.625}{0.010}\,{\tiny\textcolor{gray}{$-$0.019}} \\
 & 216 & \FIN{0.846}{0.001}\,{\tiny\textcolor{gray}{$-$0.001}} & \FIN{0.630}{0.005}\,{\tiny\textcolor{gray}{$-$0.011}} & \FIN{0.835}{0.006}\,{\tiny\textcolor{gray}{+0.000}} & \FIN{0.182}{0.003}\,{\tiny\textcolor{gray}{$-$0.035}} & \FIN{0.674}{0.019}\,{\tiny\textcolor{gray}{$-$0.051}} \\
\midrule
\multirow{2}{*}{\ORCA{}}
 & 27 & \FIN{0.848}{0.001}\,{\tiny\textcolor{gray}{$-$0.002}} & \FIN{0.689}{0.002}\,{\tiny\textcolor{gray}{$-$0.002}} & \FIN{0.888}{0.004}\,{\tiny\textcolor{gray}{$-$0.004}} & \FIN{0.573}{0.019}\,{\tiny\textcolor{gray}{$-$0.019}} & \FIN{0.762}{0.014}\,{\tiny\textcolor{gray}{$-$0.002}} \\
 & 216 & \FIN{0.849}{0.001}\,{\tiny\textcolor{gray}{$-$0.002}} & \FIN{0.713}{0.007}\,{\tiny\textcolor{gray}{+0.003}} & \FIN{0.909}{0.003}\,{\tiny\textcolor{gray}{$-$0.003}} & \FIN{0.648}{0.015}\,{\tiny\textcolor{gray}{$-$0.012}} & \FIN{0.769}{0.011}\,{\tiny\textcolor{gray}{$-$0.023}} \\
\bottomrule
\end{tabular}
\caption{Held-out test on CT-RATE/COLIPRI. Each cell is the held-out score with the held-out minus validation difference in grey. }
\label{tab:heldout}
\end{table*}

\begin{figure*}[!htb]
\centering
\includegraphics[width=\textwidth]{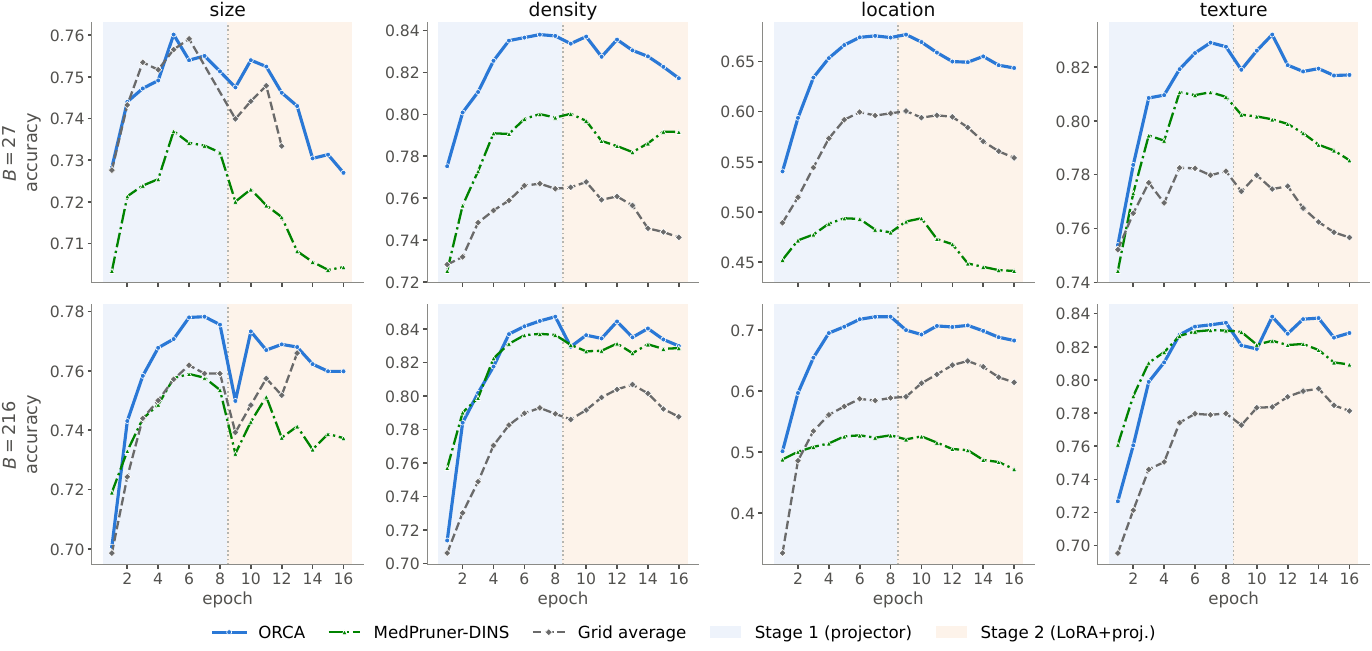}
\caption{Per-epoch VQA accuracy across the two training stages on CT-RATE/COLIPRI. Rows are token budgets, columns attribute families.}

\label{fig:vqa_2stage}
\end{figure*}

\subsection{Training curves}
\label{app:vqadyn}
Figure~\ref{fig:vqa_2stage} plots per-epoch VQA accuracy across the two training stages, s1 projector warmup then s2 LoRA, for the three compressors at both budgets. Accuracy converges by the end of s1 and s2 does not improve it, often drifting down slightly from mild overfitting, so the best epoch numbers in Table~\ref{tab:textgen} are converged rather than truncated. The ranking from the main results also holds across the whole trajectory, with ORCA's margin clearest on location and density.

\subsection{Per-attribute and per-finding scores}

Tables~\ref{tab:app_ctrate_attr_b27}--\ref{tab:app_merlin_segvol_attr} give the complete per-attribute results and Tables~\ref{tab:app_ctrate_findings}--\ref{tab:app_merlin_findings_b} the per-finding results. Per attribute, ORCA leads or ties the strongest baseline on nearly all attributes at both budgets, most clearly on location; per finding, the macro-AUROC differences across compressors are small, consistent with disease information being redundantly encoded and surviving aggressive compression.

\begin{table*}[!t]\centering\scriptsize
\begin{tblr}{
  colspec = {l Q[l,wd=2.7cm] Q[l,wd=4.0cm] c l Q[l,wd=4.3cm]},
  colsep  = 4pt,
  rowsep  = 2.5pt,
  cell{2}{1}  = {r=5}{},
  cell{7}{1}  = {r=4}{},
  cell{11}{1} = {r=3}{},
  cell{14}{1} = {r=3}{},
  row{2-6}    = {gray!12},
  row{11-13}  = {gray!12},
  hline{1,18}    = {0.08em},
  hline{2}       = {0.05em},
  hline{7,11,14} = {0.03em},
  hline{17}      = {0.05em},
}
family & attribute & gold value & probe & VQA question & what it captures \\
size & \texttt{sz\_heart\_lung} & $\log$(heart vol.\ / lung vol.) from organ masks & $R^2$ & tertile & cardiothoracic ratio; cardiomegaly proxy \\
 & \texttt{sz\_aorta\_heart} & $\log$(aorta vol.\ / heart vol.) & $R^2$ & tertile & aorto-cardiac size balance \\
 & \texttt{sz\_ivc\_aorta} & $\log$(IVC vol.\ / aorta vol.) & $R^2$ & tertile & veno-arterial caliber \\
 & \texttt{aorta\_diameter\_mm} & absolute aortic diameter in mm from the aorta mask & $R^2$ & clinical & aortic dilation / aneurysm \\
 & \texttt{heart\_width\_mm} & absolute cardiac width in mm & $R^2$ & tertile & heart size \\
density & \texttt{hu\_aorta\_calc} & fraction of aorta-wall voxels above a calcium HU cut & $R^2$ & clinical & atherosclerotic calcium burden \\
 & \texttt{vert\_median} & median vertebral-body HU & $R^2$ & tertile & bone mineral density; osteoporosis \\
 & \texttt{lung\_mean} & mean lung density in HU & $R^2$ & tertile & diffuse lung disease / fluid \\
 & \texttt{hu\_lung\_haacon} & lung high-attenuation-area fraction & $R^2$ & tertile & consolidation / ground-glass burden \\
location & \texttt{lung\_LR\_logratio} & $\log$(left-lung vol.\ / right-lung vol.) & $R^2$ & tertile & left / right lung balance \\
 & \texttt{heart\_x} & normalized left--right position of the heart centroid & $R^2$ & tertile & mediastinal laterality / shift \\
 & \texttt{ivc\_z} & normalized cranio--caudal position of the IVC centroid & $R^2$ & tertile & supero-inferior landmark height \\
texture & \texttt{lung\_fo\_Kurtosis} & lung intensity-histogram kurtosis & $R^2$ & tertile & high-density lung texture; fibrosis \\
 & \texttt{vert\_fo\_Kurtosis} & vertebral-body histogram kurtosis & $R^2$ & tertile & trabecular bone texture \\
 & \texttt{lung\_Perc15} & lung 15th-percentile HU & $R^2$ & clinical & emphysema / low-attenuation \\
disease & 18 CT-RATE / 30 Merlin & binary abnormality labels from radiology reports & AUROC & report generation & validation anchor; report-derived, not an image measurement \\
\end{tblr}
\caption{Attribute definitions and task mappings in the measurement VQA benchmark. Texture attributes are first-order statistics extracted using PyRadiomics~\cite{pyradiomics}, where fo denotes firstorder. For Merlin, the texture family is omitted because these attributes are not defined for portal-venous contrast-enhanced CT, while absolute organ HU measurements are replaced with contrast-robust inter-organ density differences.}
\label{tab:app_attr_detail}
\end{table*}

\begin{table*}[!htb]\centering\scriptsize
\begin{tblr}{
  colspec = {ll*{7}{X[c]}},
  width   = \linewidth,
  colsep  = 3pt,
  rowsep  = 1.1pt,
  cell{2}{1}  = {r=5}{},
  cell{7}{1}  = {r=4}{},
  cell{11}{1}  = {r=3}{},
  cell{14}{1}  = {r=3}{},
  row{2-6} = {gray!12},
  row{11-13} = {gray!12},
  hline{1,17}  = {0.08em},
  hline{2}    = {0.05em},
  hline{7,11,14} = {0.03em},
}
family & attribute & Grid average & Slice pooling & DivPrune & MedPruner-DINS & ToMe & MedRegion-CT & \ORCA{} \\
size & \texttt{sz\_heart\_lung} & \FIN{0.876}{0.003} & \FIN{0.844}{0.009} & \FIN{0.867}{0.003} & \FIN{0.850}{0.005} & \FIN{0.815}{0.007} & --- & \FIN{0.887}{0.004} \\
& \texttt{sz\_aorta\_heart} & \FIN{0.581}{0.016} & \FIN{0.547}{0.010} & \FIN{0.565}{0.017} & \FIN{0.546}{0.010} & \FIN{0.518}{0.007} & --- & \FIN{0.615}{0.011} \\
& \texttt{sz\_ivc\_aorta} & \FIN{0.529}{0.002} & \FIN{0.497}{0.011} & \FIN{0.499}{0.006} & \FIN{0.475}{0.009} & \FIN{0.460}{0.009} & --- & \FIN{0.541}{0.008} \\
& \texttt{aorta\_diameter\_mm} & \FIN{0.644}{0.009} & \FIN{0.636}{0.009} & \FIN{0.623}{0.015} & \FIN{0.619}{0.004} & \FIN{0.617}{0.006} & --- & \FIN{0.654}{0.002} \\
& \texttt{heart\_width\_mm} & \FIN{0.759}{0.002} & \FIN{0.730}{0.008} & \FIN{0.707}{0.004} & \FIN{0.697}{0.005} & \FIN{0.686}{0.006} & --- & \FIN{0.770}{0.006} \\
density & \texttt{hu\_aorta\_calc} & \FIN{0.838}{0.002} & \FIN{0.807}{0.012} & \FIN{0.831}{0.004} & \FIN{0.824}{0.005} & \FIN{0.807}{0.005} & --- & \FIN{0.836}{0.003} \\
& \texttt{vert\_median} & \FIN{0.540}{0.016} & \FIN{0.387}{0.003} & \FIN{0.719}{0.015} & \FIN{0.726}{0.017} & \FIN{0.566}{0.039} & --- & \FIN{0.859}{0.023} \\
& \texttt{lung\_mean} & \FIN{0.930}{0.002} & \FIN{0.909}{0.006} & \FIN{0.941}{0.003} & \FIN{0.933}{0.003} & \FIN{0.893}{0.005} & --- & \FIN{0.957}{0.001} \\
& \texttt{hu\_lung\_haacon} & \FIN{0.920}{0.004} & \FIN{0.908}{0.004} & \FIN{0.928}{0.001} & \FIN{0.919}{0.004} & \FIN{0.899}{0.004} & --- & \FIN{0.935}{0.002} \\
location & \texttt{lung\_LR\_logratio} & \FIN{0.234}{0.059} & \FIN{0.081}{0.007} & \FIN{0.247}{0.024} & \FIN{0.151}{0.005} & \FIN{0.142}{0.004} & --- & \FIN{0.684}{0.029} \\
& \texttt{heart\_x} & \FIN{0.239}{0.016} & \FIN{0.113}{0.004} & \FIN{0.128}{0.031} & \FIN{0.114}{0.004} & \FIN{0.139}{0.013} & --- & \FIN{0.629}{0.028} \\
& \texttt{ivc\_z} & \FIN{0.326}{0.009} & \FIN{0.330}{0.014} & \FIN{0.311}{0.015} & \FIN{0.268}{0.006} & \FIN{0.242}{0.005} & --- & \FIN{0.568}{0.009} \\
texture & \texttt{lung\_firstorder\_Kurtosis} & \FIN{0.918}{0.003} & \FIN{0.903}{0.005} & \FIN{0.922}{0.004} & \FIN{0.914}{0.002} & \FIN{0.890}{0.002} & --- & \FIN{0.932}{0.003} \\
& \texttt{vert\_firstorder\_Kurtosis} & \FIN{0.295}{0.011} & \FIN{0.220}{0.008} & \FIN{0.434}{0.037} & \FIN{0.339}{0.074} & \FIN{0.261}{0.009} & --- & \FIN{0.502}{0.037} \\
& \texttt{lung\_Perc15} & \FIN{0.845}{0.013} & \FIN{0.819}{0.011} & \FIN{0.876}{0.008} & \FIN{0.872}{0.006} & \FIN{0.797}{0.009} & --- & \FIN{0.899}{0.004} \\
\end{tblr}
\caption{Per-attribute probing on CT-RATE/COLIPRI at $B{=}27$, with $R^2$ for the regression families. Slice pooling uses its fixed slice count; MedRegion-CT is not applicable at $B{=}27$.}
\label{tab:app_ctrate_attr_b27}
\end{table*}

\begin{table*}[!htb]\centering\scriptsize
\begin{tblr}{
  colspec = {ll*{7}{X[c]}},
  width   = \linewidth,
  colsep  = 3pt,
  rowsep  = 1.1pt,
  cell{2}{1}  = {r=5}{},
  cell{7}{1}  = {r=4}{},
  cell{11}{1} = {r=3}{},
  cell{14}{1} = {r=3}{},
  row{2-6}    = {gray!12},
  row{11-13}  = {gray!12},
  hline{1,17}   = {0.08em},
  hline{2}      = {0.05em},
  hline{7,11,14}= {0.03em},
}
family & attribute & Grid average & Slice pooling & DivPrune & MedPruner-DINS & ToMe & MedRegion-CT & \ORCA{} \\
size & \texttt{sz\_heart\_lung} & \FIN{0.888}{0.005} & \FIN{0.844}{0.009} & \FIN{0.891}{0.005} & \FIN{0.871}{0.004} & \FIN{0.833}{0.003} & \FIN{0.905}{0.008} & \FIN{0.899}{0.001} \\
 & \texttt{sz\_aorta\_heart} & \FIN{0.606}{0.003} & \FIN{0.547}{0.010} & \FIN{0.622}{0.005} & \FIN{0.627}{0.007} & \FIN{0.543}{0.016} & \FIN{0.642}{0.013} & \FIN{0.649}{0.014} \\
 & \texttt{sz\_ivc\_aorta} & \FIN{0.525}{0.008} & \FIN{0.497}{0.011} & \FIN{0.534}{0.006} & \FIN{0.519}{0.003} & \FIN{0.483}{0.014} & \FIN{0.537}{0.002} & \FIN{0.558}{0.018} \\
 & \texttt{aorta\_diameter\_mm} & \FIN{0.653}{0.006} & \FIN{0.636}{0.009} & \FIN{0.644}{0.009} & \FIN{0.646}{0.005} & \FIN{0.638}{0.004} & \FIN{0.646}{0.004} & \FIN{0.675}{0.011} \\
 & \texttt{heart\_width\_mm} & \FIN{0.755}{0.002} & \FIN{0.730}{0.008} & \FIN{0.734}{0.004} & \FIN{0.717}{0.017} & \FIN{0.706}{0.008} & \FIN{0.744}{0.002} & \FIN{0.822}{0.001} \\
density & \texttt{hu\_aorta\_calc} & \FIN{0.843}{0.009} & \FIN{0.807}{0.012} & \FIN{0.851}{0.007} & \FIN{0.863}{0.006} & \FIN{0.835}{0.003} & \FIN{0.853}{0.005} & \FIN{0.868}{0.005} \\
 & \texttt{vert\_median} & \FIN{0.736}{0.020} & \FIN{0.387}{0.003} & \FIN{0.814}{0.010} & \FIN{0.871}{0.007} & \FIN{0.724}{0.015} & \FIN{0.870}{0.012} & \FIN{0.884}{0.014} \\
 & \texttt{lung\_mean} & \FIN{0.952}{0.002} & \FIN{0.909}{0.006} & \FIN{0.954}{0.003} & \FIN{0.949}{0.002} & \FIN{0.917}{0.006} & \FIN{0.959}{0.004} & \FIN{0.962}{0.002} \\
 & \texttt{hu\_lung\_haacon} & \FIN{0.936}{0.003} & \FIN{0.908}{0.004} & \FIN{0.935}{0.006} & \FIN{0.933}{0.002} & \FIN{0.914}{0.004} & \FIN{0.933}{0.003} & \FIN{0.941}{0.001} \\
location & \texttt{lung\_LR\_logratio} & \FIN{0.228}{0.043} & \FIN{0.081}{0.007} & \FIN{0.253}{0.029} & \FIN{0.207}{0.063} & \FIN{0.190}{0.063} & \FIN{0.274}{0.067} & \FIN{0.735}{0.019} \\
 & \texttt{heart\_x} & \FIN{0.168}{0.009} & \FIN{0.113}{0.004} & \FIN{0.164}{0.014} & \FIN{0.169}{0.010} & \FIN{0.185}{0.013} & \FIN{0.166}{0.005} & \FIN{0.687}{0.019} \\
 & \texttt{ivc\_z} & \FIN{0.368}{0.009} & \FIN{0.330}{0.014} & \FIN{0.368}{0.010} & \FIN{0.342}{0.012} & \FIN{0.291}{0.006} & \FIN{0.377}{0.006} & \FIN{0.611}{0.014} \\
texture & \texttt{lung\_firstorder\_Kurtosis} & \FIN{0.929}{0.005} & \FIN{0.903}{0.005} & \FIN{0.931}{0.004} & \FIN{0.928}{0.004} & \FIN{0.908}{0.005} & \FIN{0.936}{0.004} & \FIN{0.933}{0.002} \\
 & \texttt{vert\_firstorder\_Kurtosis} & \FIN{0.473}{0.075} & \FIN{0.220}{0.008} & \FIN{0.433}{0.054} & \FIN{0.477}{0.054} & \FIN{0.398}{0.056} & \FIN{0.581}{0.026} & \FIN{0.600}{0.007} \\
 & \texttt{lung\_Perc15} & \FIN{0.889}{0.001} & \FIN{0.819}{0.011} & \FIN{0.900}{0.004} & \FIN{0.905}{0.002} & \FIN{0.841}{0.009} & \FIN{0.916}{0.002} & \FIN{0.915}{0.003} \\
\end{tblr}
\caption{Per-attribute probing on CT-RATE/COLIPRI at $B{=}216$, with $R^2$ for the regression families. MedRegion-CT is shown at its fixed $\bar{N}{=}549$ and Slice pooling at its fixed slice count.}
\label{tab:app_ctrate_attr_b216}
\end{table*}

\begin{table*}[!h]\centering\scriptsize
\begin{tblr}{
  colspec = {ll*{7}{X[c]}},
  width   = \linewidth,
  colsep  = 3.5pt,
  rowsep  = 1.1pt,
  cell{1}{1} = {r=2}{},
  cell{1}{2} = {r=2}{},
  cell{1}{3} = {c=3}{c},
  cell{1}{6} = {c=4}{c},
  cell{3}{1} = {r=6}{},
  cell{9}{1} = {r=3}{},
  cell{12}{1}= {r=3}{},
  row{3-8}   = {gray!12},
  row{12-14} = {gray!12},
  vline{6}   = {0.03em},
  hline{1,15}= {0.08em},
  hline{2}   = {3-9}{0.03em},
  hline{3}   = {0.05em},
}
family & attribute & SuPreM $B{=}216$ & & & SegVol $B{=}256$ & & & \\
 & & Grid average & ToMe & \ORCA{} & Grid average & MedPruner-DINS & ToMe & \ORCA{} \\
size & \texttt{spleen\_ml} & \FIN{0.717}{0.020} & \FIN{0.510}{0.023} & \FIN{0.802}{0.005} & \FIN{0.771}{0.022} & \FIN{0.811}{0.023} & \FIN{0.621}{0.145} & \FIN{0.828}{0.002} \\
 & \texttt{spleen\_vert\_ratio} & \FIN{0.651}{0.024} & \FIN{0.478}{0.017} & \FIN{0.722}{0.007} & \FIN{0.658}{0.030} & \FIN{0.731}{0.021} & \FIN{0.522}{0.105} & \FIN{0.733}{0.014} \\
 & \texttt{kidney\_ml} & \FIN{0.405}{0.006} & \FIN{0.334}{0.012} & \FIN{0.500}{0.026} & \FIN{0.400}{0.065} & \FIN{0.538}{0.009} & \FIN{0.343}{0.030} & \FIN{0.552}{0.015} \\
 & \texttt{kidney\_vert\_ratio} & \FIN{0.395}{0.019} & \FIN{0.338}{0.007} & \FIN{0.501}{0.005} & \FIN{0.306}{0.040} & \FIN{0.511}{0.031} & \FIN{0.326}{0.047} & \FIN{0.551}{0.011} \\
 & \texttt{aorta\_diameter\_mm} & \FIN{0.424}{0.004} & \FIN{0.434}{0.004} & \FIN{0.506}{0.027} & \FIN{0.494}{0.010} & \FIN{0.538}{0.038} & \FIN{0.409}{0.018} & \FIN{0.561}{0.010} \\
 & \texttt{aorta\_vert\_diam} & \FIN{0.181}{0.001} & \FIN{0.195}{0.004} & \FIN{0.256}{0.016} & \FIN{0.192}{0.010} & \FIN{0.279}{0.044} & \FIN{0.133}{0.007} & \FIN{0.322}{0.008} \\
density & \texttt{vert\_L1T12\_median} & \FIN{0.707}{0.010} & \FIN{0.708}{0.008} & \FIN{0.794}{0.008} & \FIN{0.768}{0.002} & \FIN{0.763}{0.035} & \FIN{0.767}{0.021} & \FIN{0.807}{0.004} \\
 & \texttt{muscle\_auto\_median} & \FIN{0.914}{0.005} & \FIN{0.864}{0.009} & \FIN{0.954}{0.003} & \FIN{0.788}{0.020} & \FIN{0.775}{0.040} & \FIN{0.758}{0.004} & \FIN{0.821}{0.020} \\
 & \texttt{liver\_spleen\_diff} & \FIN{0.427}{0.035} & \FIN{0.373}{0.023} & \FIN{0.644}{0.038} & \FIN{0.336}{0.025} & \FIN{0.477}{0.027} & \FIN{0.398}{0.017} & \FIN{0.447}{0.037} \\
location & \texttt{kidney\_z} & \FIN{0.495}{0.009} & \FIN{0.457}{0.007} & \FIN{0.675}{0.015} & \FIN{0.581}{0.026} & \FIN{0.718}{0.018} & \FIN{0.640}{0.027} & \FIN{0.746}{0.017} \\
 & \texttt{kidney\_lr\_z\_asym} & \FIN{0.305}{0.030} & \FIN{0.173}{0.005} & \FIN{0.689}{0.010} & \FIN{0.419}{0.019} & \FIN{0.753}{0.013} & \FIN{0.501}{0.091} & \FIN{0.789}{0.005} \\
 & \texttt{bladder\_z} & \FIN{0.385}{0.022} & \FIN{0.257}{0.006} & \FIN{0.437}{0.036} & \FIN{0.166}{0.016} & \FIN{0.209}{0.023} & \FIN{0.208}{0.015} & \FIN{0.284}{0.075} \\
\end{tblr}
\caption{Per-attribute probing on Merlin, SuPreM at $B{=}216$ and SegVol at $B{=}256$, as mean $R^2 \pm$ standard deviation over three seeds. MedPruner-DINS applies only to SegVol, since SuPreM's windowed-attention backbone gives no per-token saliency; Merlin has no texture family.}
\label{tab:app_merlin_suprem_attr}
\label{tab:app_merlin_segvol_attr}
\end{table*}

\begin{table*}[!htb]\centering\scriptsize
\begin{tblr}{
  colspec = {l*{7}{X[c]}},
  width   = \linewidth,
  colsep  = 2.5pt,
  rowsep  = 1.1pt,
  hline{1,20} = {0.08em},
  hline{2}    = {0.05em},
}
finding & Avg.\ pool & Slice pool & DivPrune & MedPruner & ToMe & MedRegion & \ORCA{} \\
\texttt{Medical material} & \FIN{0.932}{0.001} & \FIN{0.930}{0.003} & \FIN{0.932}{0.003} & \FIN{0.932}{0.002} & \FIN{0.929}{0.002} & \FIN{0.932}{0.001} & \FIN{0.932}{0.001} \\
\texttt{Arterial wall calcification} & \FIN{0.933}{0.001} & \FIN{0.931}{0.001} & \FIN{0.932}{0.001} & \FIN{0.933}{0.001} & \FIN{0.932}{0.001} & \FIN{0.930}{0.000} & \FIN{0.932}{0.000} \\
\texttt{Cardiomegaly} & \FIN{0.932}{0.002} & \FIN{0.932}{0.001} & \FIN{0.931}{0.001} & \FIN{0.930}{0.002} & \FIN{0.927}{0.001} & \FIN{0.933}{0.002} & \FIN{0.934}{0.001} \\
\texttt{Pericardial effusion} & \FIN{0.862}{0.006} & \FIN{0.861}{0.004} & \FIN{0.866}{0.005} & \FIN{0.864}{0.000} & \FIN{0.861}{0.007} & \FIN{0.860}{0.004} & \FIN{0.866}{0.005} \\
\texttt{Coronary artery wall calcification} & \FIN{0.935}{0.001} & \FIN{0.935}{0.001} & \FIN{0.936}{0.001} & \FIN{0.937}{0.000} & \FIN{0.937}{0.001} & \FIN{0.935}{0.002} & \FIN{0.936}{0.001} \\
\texttt{Hiatal hernia} & \FIN{0.718}{0.002} & \FIN{0.715}{0.001} & \FIN{0.717}{0.002} & \FIN{0.712}{0.004} & \FIN{0.713}{0.003} & \FIN{0.713}{0.002} & \FIN{0.716}{0.001} \\
\texttt{Lymphadenopathy} & \FIN{0.757}{0.005} & \FIN{0.755}{0.001} & \FIN{0.759}{0.003} & \FIN{0.762}{0.001} & \FIN{0.758}{0.002} & \FIN{0.756}{0.003} & \FIN{0.759}{0.002} \\
\texttt{Emphysema} & \FIN{0.813}{0.006} & \FIN{0.809}{0.004} & \FIN{0.811}{0.004} & \FIN{0.812}{0.005} & \FIN{0.809}{0.003} & \FIN{0.813}{0.004} & \FIN{0.814}{0.004} \\
\texttt{Atelectasis} & \FIN{0.818}{0.002} & \FIN{0.814}{0.001} & \FIN{0.816}{0.003} & \FIN{0.818}{0.006} & \FIN{0.813}{0.005} & \FIN{0.819}{0.004} & \FIN{0.822}{0.002} \\
\texttt{Lung nodule} & \FIN{0.743}{0.003} & \FIN{0.742}{0.003} & \FIN{0.743}{0.001} & \FIN{0.744}{0.005} & \FIN{0.740}{0.001} & \FIN{0.738}{0.001} & \FIN{0.743}{0.002} \\
\texttt{Lung opacity} & \FIN{0.873}{0.002} & \FIN{0.872}{0.001} & \FIN{0.872}{0.002} & \FIN{0.874}{0.001} & \FIN{0.869}{0.001} & \FIN{0.871}{0.000} & \FIN{0.872}{0.002} \\
\texttt{Pulmonary fibrotic sequela} & \FIN{0.752}{0.006} & \FIN{0.745}{0.008} & \FIN{0.752}{0.006} & \FIN{0.753}{0.005} & \FIN{0.747}{0.004} & \FIN{0.751}{0.002} & \FIN{0.751}{0.002} \\
\texttt{Pleural effusion} & \FIN{0.969}{0.003} & \FIN{0.969}{0.002} & \FIN{0.970}{0.002} & \FIN{0.967}{0.000} & \FIN{0.968}{0.002} & \FIN{0.969}{0.001} & \FIN{0.968}{0.001} \\
\texttt{Mosaic attenuation pattern} & \FIN{0.892}{0.002} & \FIN{0.893}{0.004} & \FIN{0.894}{0.000} & \FIN{0.891}{0.003} & \FIN{0.880}{0.003} & \FIN{0.890}{0.003} & \FIN{0.893}{0.002} \\
\texttt{Peribronchial thickening} & \FIN{0.814}{0.001} & \FIN{0.808}{0.002} & \FIN{0.811}{0.002} & \FIN{0.812}{0.002} & \FIN{0.805}{0.001} & \FIN{0.814}{0.008} & \FIN{0.817}{0.001} \\
\texttt{Consolidation} & \FIN{0.924}{0.002} & \FIN{0.925}{0.001} & \FIN{0.925}{0.002} & \FIN{0.922}{0.001} & \FIN{0.922}{0.002} & \FIN{0.924}{0.002} & \FIN{0.925}{0.002} \\
\texttt{Bronchiectasis} & \FIN{0.809}{0.005} & \FIN{0.800}{0.005} & \FIN{0.808}{0.005} & \FIN{0.807}{0.007} & \FIN{0.797}{0.002} & \FIN{0.809}{0.008} & \FIN{0.807}{0.012} \\
\texttt{Interlobular septal thickening} & \FIN{0.886}{0.002} & \FIN{0.886}{0.003} & \FIN{0.884}{0.001} & \FIN{0.886}{0.001} & \FIN{0.879}{0.003} & \FIN{0.885}{0.007} & \FIN{0.883}{0.002} \\
\end{tblr}
\captionsetup{justification=raggedright,singlelinecheck=false}
\caption{Per-finding probing on CT-RATE at $B{=}216$, macro-AUROC.}
\label{tab:app_ctrate_findings}
\end{table*}

\begin{table}[!htb]\centering\scriptsize
\begin{tblr}{
  colspec = {lcc},
  colsep  = 4pt,
  rowsep  = 1.1pt,
  hline{1,32} = {0.08em},
  hline{2}    = {0.05em},
}
finding & Avg.\ pool & \ORCA{} \\
\texttt{submucosal\_edema} & \FIN{0.746}{0.005} & \FIN{0.760}{0.005} \\
\texttt{renal\_hypodensities} & \FIN{0.722}{0.004} & \FIN{0.717}{0.004} \\
\texttt{aortic\_valve\_calcification} & \FIN{0.882}{0.004} & \FIN{0.880}{0.006} \\
\texttt{coronary\_calcification} & \FIN{0.807}{0.006} & \FIN{0.827}{0.012} \\
\texttt{thrombosis} & \FIN{0.710}{0.004} & \FIN{0.719}{0.022} \\
\texttt{metastatic\_disease} & \FIN{0.818}{0.005} & \FIN{0.779}{0.009} \\
\texttt{pancreatic\_atrophy} & \FIN{0.808}{0.004} & \FIN{0.819}{0.005} \\
\texttt{renal\_cyst} & \FIN{0.714}{0.001} & \FIN{0.717}{0.002} \\
\texttt{osteopenia} & \FIN{0.953}{0.001} & \FIN{0.944}{0.002} \\
\texttt{surgically\_absent\_gallbladder} & \FIN{0.723}{0.003} & \FIN{0.717}{0.005} \\
\texttt{atelectasis} & \FIN{0.792}{0.002} & \FIN{0.793}{0.014} \\
\texttt{abdominal\_aortic\_aneurysm} & \FIN{0.853}{0.015} & \FIN{0.897}{0.014} \\
\texttt{anasarca} & \FIN{0.969}{0.001} & \FIN{0.968}{0.001} \\
\texttt{hiatal\_hernia} & \FIN{0.796}{0.001} & \FIN{0.799}{0.004} \\
\texttt{lymphadenopathy} & \FIN{0.625}{0.010} & \FIN{0.683}{0.031} \\
\texttt{prostatomegaly} & \FIN{0.764}{0.007} & \FIN{0.774}{0.009} \\
\texttt{biliary\_ductal\_dilation} & \FIN{0.782}{0.004} & \FIN{0.759}{0.016} \\
\texttt{cardiomegaly} & \FIN{0.817}{0.003} & \FIN{0.820}{0.010} \\
\texttt{splenomegaly} & \FIN{0.787}{0.006} & \FIN{0.882}{0.006} \\
\texttt{hepatomegaly} & \FIN{0.813}{0.016} & \FIN{0.808}{0.027} \\
\texttt{atherosclerosis} & \FIN{0.816}{0.001} & \FIN{0.825}{0.004} \\
\texttt{ascites} & \FIN{0.947}{0.002} & \FIN{0.939}{0.005} \\
\texttt{pleural\_effusion} & \FIN{0.858}{0.003} & \FIN{0.864}{0.010} \\
\texttt{hepatic\_steatosis} & \FIN{0.822}{0.009} & \FIN{0.807}{0.055} \\
\texttt{appendicitis} & \FIN{0.625}{0.038} & \FIN{0.600}{0.026} \\
\texttt{gallstones} & \FIN{0.720}{0.005} & \FIN{0.707}{0.003} \\
\texttt{hydronephrosis} & \FIN{0.705}{0.009} & \FIN{0.723}{0.004} \\
\texttt{bowel\_obstruction} & \FIN{0.800}{0.016} & \FIN{0.839}{0.017} \\
\texttt{free\_air} & \FIN{0.814}{0.002} & \FIN{0.832}{0.008} \\
\texttt{fracture} & \FIN{0.768}{0.002} & \FIN{0.757}{0.003} \\
\end{tblr}
\caption{Per-finding probing on Merlin/SuPreM at $B{=}216$, macro-AUROC over all 30 findings.}
\label{tab:app_merlin_findings_b}
\end{table}

\end{document}